\documentclass{article}

\usepackage[preprint, nonatbib]{neurips_2021}

\usepackage[numbers]{natbib}
\usepackage[utf8]{inputenc} 
\usepackage[T1]{fontenc}    
\usepackage{hyperref}       
\usepackage{url}            
\usepackage{booktabs}       
\usepackage{amsfonts}       
\usepackage{nicefrac}       
\usepackage{microtype}      
\usepackage{xcolor}         
\usepackage{todonotes}
\usepackage{bm}
\usepackage{graphicx}
\usepackage{subcaption}

\usepackage{siunitx} 
\usepackage{caption}
\usepackage{amsmath} 
\usepackage{amssymb}
\usepackage{listings} 
\usepackage{hhline} 

\title{A Loss Curvature Perspective on Training Instability in Deep Learning}

\author{%
Justin Gilmer \thanks{Equal Contribution.} \\\texttt{gilmer@google.com}\And
Behrooz Ghorbani \footnotemark[1] \\\texttt{ghorbani@google.com}\And
Ankush Garg\\\texttt{ankugarg@google.com}\AND
Sneha Kudugunta\\\texttt{snehakudugunta@google.com}\And
Behnam Neyshabur\\\texttt{neyshabur@google.com}\And
David Cardoze\\\texttt{dcardoze@google.com}\And
George Dahl\\\texttt{gdahl@google.com}\And
Zack Nado\\\texttt{znado@google.com}\And
Orhan Firat\\\texttt{orhanf@google.com}\\~
}

\begin{document}

\maketitle

\begin{abstract}
 In this work, we study the evolution of the loss Hessian across many classification tasks in order to understand the effect the curvature of the loss has on the training dynamics. Whereas prior work has focused on how different learning rates affect the loss Hessian observed during training, we also analyze the effects of model initialization, architectural choices, and common training heuristics such as gradient clipping and learning rate warmup. Our results demonstrate that successful model and hyperparameter choices allow the early optimization trajectory to either avoid---or navigate out of---regions of high curvature and into flatter regions that tolerate a higher learning rate. Our results suggest a unifying perspective on how disparate mitigation strategies for training instability ultimately address the same underlying failure mode of neural network optimization, namely poor conditioning. Inspired by the conditioning perspective, we show that learning rate warmup can improve training stability just as much as batch normalization, layer normalization, MetaInit, GradInit, and Fixup initialization.
\end{abstract}

\section{Introduction}

Optimization of neural networks can easily fail. While recent architectural advances such as skip connections \citep{he2016deep} and Batch Normalization \citep{ioffe2015batch} have been applied successfully to produce architectures and hyperparameters that reliably train well, even small changes to a trainable configuration can easily result in training that diverges. More generally, producing a configuration that strikes the right balance between stable training and rapid optimization progress on a new domain can be difficult---practitioners and researchers have few reliable heuristics to guide them through the process. As a result, the specific hyperparameter tuning protocol has an outsized influence on the results \citep{choi2019empirical,pmlr-v119-sivaprasad20a} and successes often rely on large hyperparameter searches \citep{NadoEtAl2021_largeBatchReality}. Developing a principled understanding of what makes general architectures trainable would allow researchers to more reliably navigate this process and has the potential to dramatically accelerate research into finding better, more scalable architectures.

\begin{figure}
     \centering

    \includegraphics[width=\textwidth]{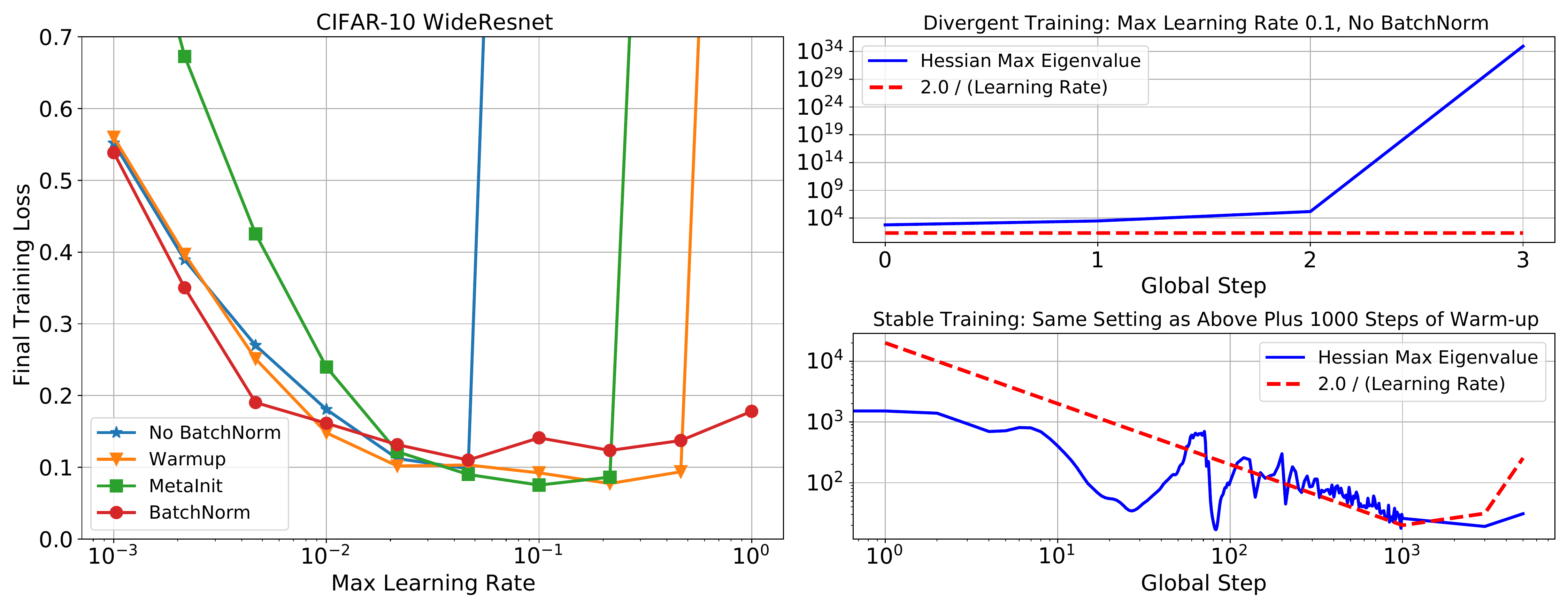}

     \caption{Left: Three different methods which when applied to a WideResnet 28-10 architecture (w/o Batch Normalization) enable training at larger learning rates: learning rate warmup,  MetaInit, and adding normalization layers. Each point reports the final training loss after training with cosine decay for 300 epochs. The test performance of the models closely matches the training loss behavior (see Figure~\ref{fig:wideresnet_test_performance} of the appendix). Right: The evolution of the largest eigenvalue of the Hessian throughout the training for the No-BatchNorm model with and without warm-up.}
     \label{fig:wideresnet_performance}
\end{figure}

The focus of the empirical investigation of this work is to better understand what limits the maximum trainable learning rate for deep learning models trained with the typical minibatch stochastic gradient descent (SGD) family algorithms. As part of this investigation, we examine several methods developed by the deep learning community that have enabled training at larger learning rates and improved performance. Many methods have been developed that can achieve this goal, notably normalization, learning rate warmup, gradient clipping \citep{pascanu2013difficulty}, and better model initializations such as Fixup \citep{zhang2019fixup}, MetaInit \citep{dauphin2019metainit}, and GradInit \citep{zhu2021gradinit}. While these methods are certainly not exactly equivalent, a key property they all have in common is that they can enable training at larger learning rates when applied to certain models (see for example Figure~\ref{fig:wideresnet_performance}). 

A natural hypothesis is that methods which enable training at larger learning rates do so through reducing the \emph{sharpness}\footnote{Throughout this work we will use the term sharpness to refer to the maximum eigenvalue of the loss Hessian, denoted as $\lambda_1$.} of the loss surface during training. Indeed, this hypothesis has already been proposed as one of the beneficial effects of Batch Normalization \citep{ghorbani2019investigation, santurkar2018does} and residual connections \citep{li2017visualizing}, and quadratic models of the loss surface predict that optimization with SGD is unstable when $\lambda_1 > 2 / \eta$ \cite{wu2018sgd}. However, recent empirical investigations into the relevance of quadratic stability bounds to neural network training have either focused on smaller models, focused on full batch training at small learning rates, and do not investigate connections between sharpness, model initialization and learning rate warmup \citep{cohen2021gradient, jastrzebski2020break}.

In this work, we design a series of large scale experiments studying the evolution of the loss sharpness as we vary the learning rate, warmup period, initialization, and architectural choices. Our results demonstrate the central role that $\lambda_1$ plays in neural network optimization—maintaining sufficiently small $\lambda_1$ during optimization is a necessary condition for successful training at large learning rates. Consequently, reducing $\lambda_1$ is a primary benefit of proper tuning of a number of architecture and optimization hyperparameters: including model initialization, location of normalization, and warmup schedule. Specifically, we show the following:

\begin{itemize}
    \item We provide large scale empirical confirmation that training of neural networks with SGD+momentum is stable only when the optimization trajectory primarily resides in a region of parameter space where $\lambda_1 \lesssim 2 / \eta$, where $\eta$ denotes the learning rate. This corroborates the theoretical predictions in \citet{wu2018sgd} and recent empirical observations of \citet{jastrzebski2020break} and \citet{cohen2021gradient}.
    \item We demonstrate that several successful initialization strategies for architectures without normalization operate primarily by reducing curvature early in training, enabling training at larger learning rates.
    \item We show that learning rate warmup gradually reduces $\lambda_1$ during training, offering similar benefits to better model initialization. We connect the mechanism by which warmup operates to the dynamical stability model of \citet{wu2018sgd}.
    \item We show that learning rate warmup is a simple yet competitive baseline for research into better model initialization. We demonstrate that key progress in this area from \citet{dauphin2019metainit, zhang2019fixup, zhu2021gradinit} can be matched by the application of learning rate warmup alone.
    \item Finally, we show that large loss curvature can result in poor scaling at large batch sizes and interventions designed to improve loss conditioning can drastically improve the model's ability to leverage data parallelism.
\end{itemize}

\section{Related Work}

{\bf Understanding BatchNorm} The loss Hessian has been a central object of study for understanding optimization of neural networks. \citet{santurkar2018does} argues that an important benefit of Batch Normalization is improved smoothness of the loss surface, while \citet{lewkowycz2020large} notes that this is improved smoothness is only observed when higher learning rates are used in combination with Batch Normalization. Our results are generally consistent with this current understanding of Batch Normalization, however some of our experiments provide additional nuance---notably we observe several instances where models suffer from training instability (and high loss curvature) early in training despite using Batch Normalization (see Section~\ref{sec-curvature}).

{\bf Evolution of the loss Hessian} Recent research has closely studied the interaction between sharpness and learning rate. \citet{cohen2021gradient} observed that for full batch training, the loss curvature demonstrates ``progressive sharpening'', where it increases until $\lambda_1 \approx 2.0 / \eta$. In our experiments we observe that progressive sharpening of the loss occurs for models trained with SGD, with a notable exception of Transformer models on LM1B. \citet{lewkowycz2020large} provides a theoretical model predicting that some architectures initialized at point with $\lambda_1 > 2.0 / \eta$ and trained with an MSE loss may enter a ``catapult'' regime---where the loss increases early until a flatter region of the loss surface is found, with divergence occurring in cases where $\lambda_1$ greatly exceeds $2.0 / \eta$. Our experimental setup violates many assumptions from this work, though it is noteworthy that some of the behavior we observe is consistent with catapult dynamics---we observed that models can consistently be trained even when $\lambda_1 > 2.0 / \eta$ at initialization, resulting in brief initial training instability until optimization enters a flatter region. However, we observe some cases where models diverge despite being initialized at a point where $\lambda_1 \leq 2.0 / \eta$, which is not predicted by this theoretical model.

 \citet{jastrzebski2020break} describes several preconditioning benefits of using large learning rates with SGD, namely smaller observed $\lambda_1$ and larger $\lambda_k / \lambda_1$ during training. Our results provide large scale empirical evidence for the conjecture that larger learning rates result in flatter curvature during training. Additionally, we demonstrate that learning rate warmup and gradient clipping leverage this mechanism to reduce $\lambda_1$ even further, as some models can be trained with even larger learning rates when these two techniques are adopted.
 
\section{Experimental Setup}
We investigate models trained on several benchmarks: CIFAR-10 \citep{cifarKrizhevsky09learningmultiple} and ImageNet \citep{russakovsky2015imagenet} for image classification, LM1B \citep{lm1b} for Language Modeling, and WMT for Neural Machine Translation (NMT). On CIFAR-10 we consider the WideResnet \citep{zagoruyko2016wide} and DenseNet \citep{huang2017densely} architectures, both with and without Batch Normalization. When training without Batch Normalization we consider several initialization strategies including the default ``LeCun Normal'' initialization, and running MetaInit. As a way to artificially induce worse initializations, we also consider experiments where we scale every variable produced by the default initialization by a constant factor $\alpha$. The NMT models are trained on the WMT'16 EN-DE training set, tuned for hyper-parameters on the WMT'16 EN-DE validation set and evaluated on the WMT'14 EN-DE test set for BLEU scores. For NMT and LM1B Language Modeling, we train 6 layer Transformer models \citep{vaswani_transformers}. Inspired from \citet{layernormstransformer}, we experiment with three Layer Norm settings: pre-Layer Norm, post-Layer Norm \citep{liu2020understanding} and no Layer Norm for the transformer models.

Each model is trained with various learning rates using cosine decay (unless mentioned explicitly). For warmup experiments we use linear warmup which starts at 0 and scales linearly to a max value $\eta$ before applying cosine decay. To measure the max eigenvalue of the loss Hessian we use the Lanczos method where the number of iterations varied as needed depending on the architecture (details provided in the appendix). 

\section{Early Training Instability and the Loss Hessian}
\label{sec-curvature}

\begin{figure}
    \centering
    \includegraphics[width=.95\textwidth]{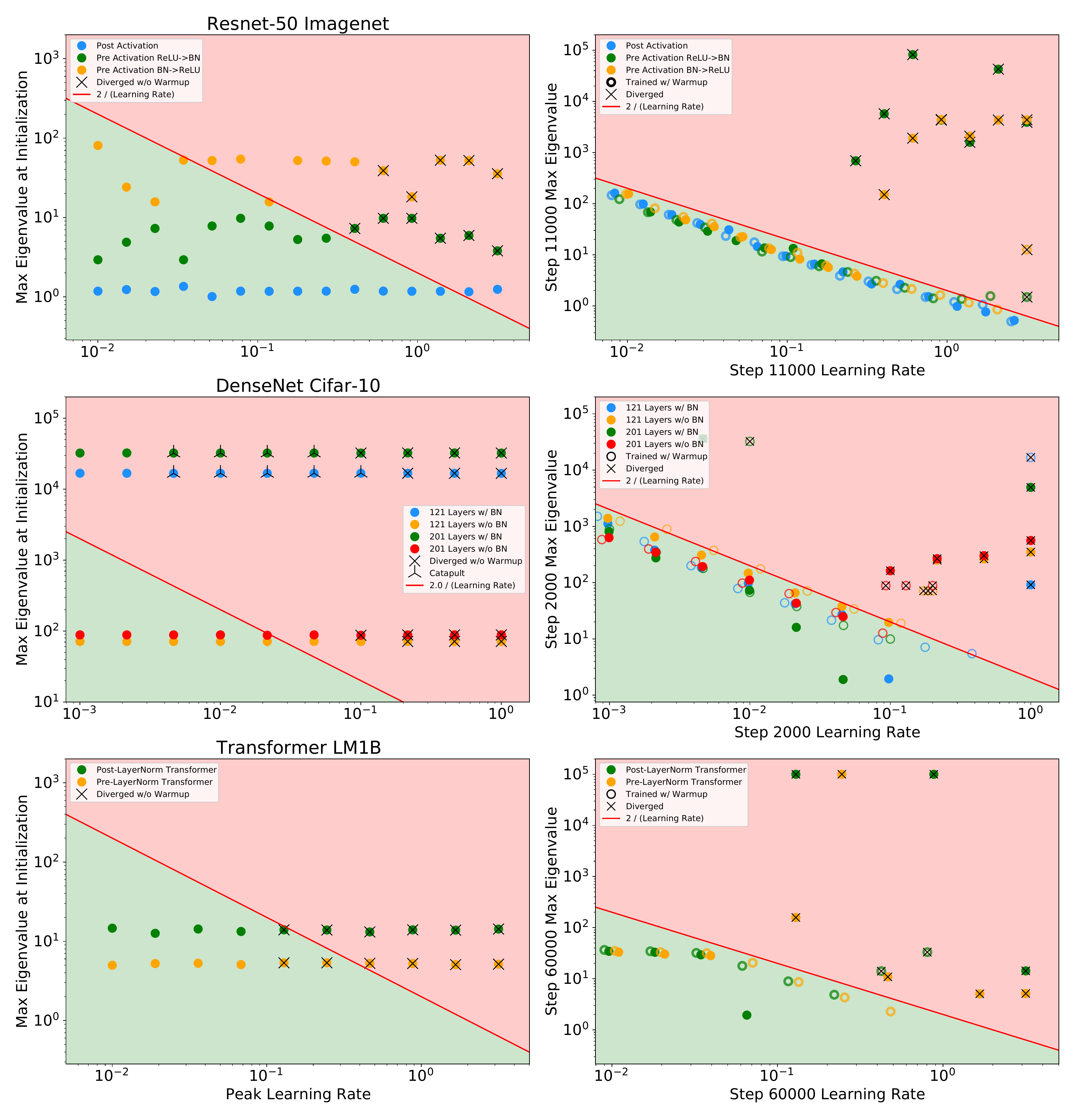}
    \caption{Measurements of the Hessian max eigenvalue of different models at initialization (left) and during training (right). Divergent models are indicated with an X. On the left, we plot $\lambda_1$ at initialization along with the peak learning used during training. On the right, we plot $\eta$ and $\lambda_1$ at a specified step in training. For divergent models on the right plot, we record the last learning rate and max eigenvalue that occur before divergence is detected (defined as observing a NaN in the loss). Across all datasets and models, successful training occurs only when optimization enters a stable region of parameter space where $\lambda_1 \leq 2 / \eta$. Models with higher initial loss curvature tend to diverge at smaller learning rates relative to models initialized with flatter curvature, however larger learning rates are possible in these cases when warmup is used. Note, to avoid overlapping points we have applied a small deterministic shift to the x-position of points on the right hand plots. In the case of DenseNets w/ BN we observed some models exhibit catapult behavior (loss increases early followed by normal training), these have been marked accordingly.}
    \label{fig:main_plot}
\end{figure}

In Figure~\ref{fig:main_plot} we plot the curvature at initialization and during training for a series of models trained on different datasets (plots showing final performance for all models can be found in the appendix). Each row indicates a different base model, the left column plots the curvature of the model at initialization and indicates with an `X' whether or not the model diverges when trained without warmup. On the right we plot the measured curvature and learning rate at a specified point during training. We observe across all datasets that successful training occurs only when optimization enters a region of parameter space where $\lambda_1 \leq 2 / \eta$, and that divergent models are outside this region shortly before divergence. At initialization, some models can be successfully trained even when they start out in the unstable region and generally speaking, divergence is more likely for models deeper in the unstable region.

These experiments also visualize how different methods unlock training at higher learning rates. For CIFAR-10 WideResnet, removing batch norm results in a model with higher curvature at initialization and results in divergent models when trained with a learning rate $\eta > .1$. Scaling the WideResnet initialization up by a factor of $1.5$ exacerbates the problem, resulting in even higher curvature at initialization and divergence when $\eta > 10^{-2}$. MetaInit starts the model out at a point with very small $\lambda_1$, and allows training without Batch Normalization at higher learning rates than the default initialization. We also observed that higher learning rates can be unlocked when the models are trained with learning rate warmup, and that we are able to match the performance of MetaInit using warmup and starting with the bad initialization. Warmup was particularly effective for models which exhibit large $\lambda_1$ either at initialization or early in training. Other models such as the post activation Resnet-50, and WideResnet w/ Batch Normalization did not benefit from warmup at the considered learning rates.

For the DenseNet experiments, it is noteworthy that the models with Batch Normalization actually start out with higher curvature than the non-BN variants. This is contrary to the generally accepted narrative that Batch Normalization improves the smoothness of the loss surface \citep{ghorbani2019investigation, santurkar2018does}, though not without precedent as \citet{yang2019mean} demonstrates settings for which Batch Normalization can cause exploding gradients. We found that the Batch Normalization models were more unstable than the non-BN variants here, as some models diverged at smaller learning rates. However, when combined with warmup the BN models were trainable at learning rates $\eta > .1$, whereas this did not hold for the non-BN variants, which diverge both with and without warmup at these learning rates. This result suggests that BN still offers training stability for this model, and flatter curvature mid training \emph{if trained with warmup and a higher learning rate}, however no smoothness benefits are observed at initialization.

For Resnet-50 trained on ImageNet we compare two different residual blocks: the preactivation block \citep{he2016identity} and the more commonly used post activation block \citep{he2016deep}. For the preactivation block, we also consider flipping the order of the ReLU activation and batch normalization, as was considered in \citet{brock2021characterizing}. We find that both preactivation models start out in a region of higher curvature relative to the post activation variant, and that these models diverge when $\eta > .5$ whereas the post activation variant is trainable with learning rates as large at $10$. 

Notably, there are several models in our experiments which diverge despite starting out in a region where $\lambda_1 < 2 / \eta$. This occurs for both the pre and post layernorm transformer, and the WideResnet model initialized with MetaInit. We found for these divergent models that the curvature rapidly increases in the initial steps of training, which is partially visible in the mid training plot where we plot the final observed curvature before divergence. Full training curves for these models can be found in the appendix. These examples are noteworthy because it demonstrates that measuring smoothness at model initialization is not always sufficient to predict whether or not the model will be easily trained. Currently, some architectural innovations are motivated by an analysis of either gradient statistics or smoothness at initialization \citep{liu2020understanding}---a more robust analysis would consider the evolution of these statistics under SGD.

\begin{figure}
     \centering
     \includegraphics[width=\textwidth]{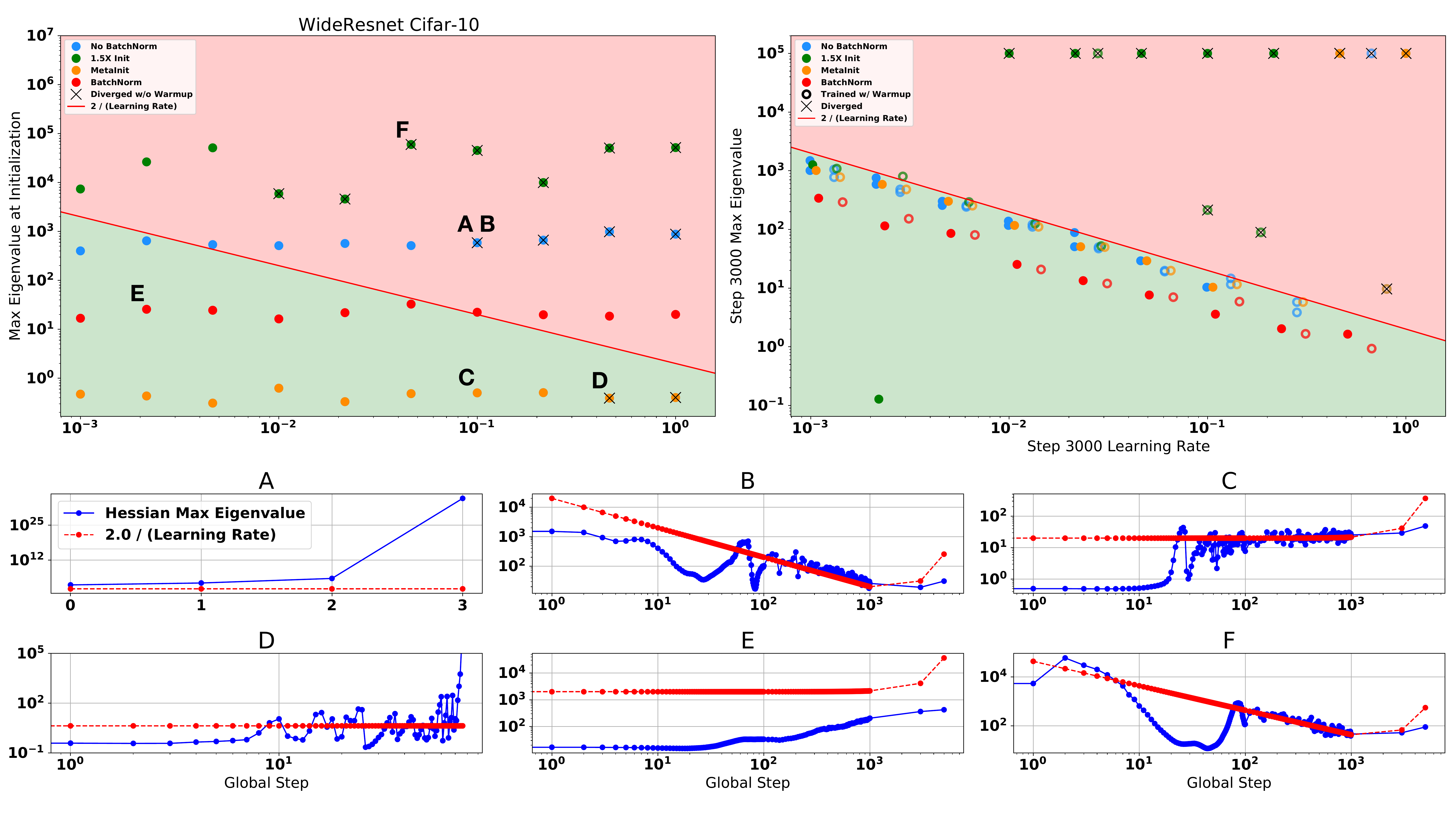}
     \caption{ {\bf Top row:} Measured curvature at initialization (left) and mid training (right) for variants of the WideResnet 28-10 model on Cifar10. Models which train successfully enter a region where $\lambda_1 \leq 2.0 / \eta$ mid training (or fluctuate just above this bound). Both batch norm and the MetaInit initialization reduce the curvature early in training, allowing training at larger learning rates. For models with high initial curvature, learning rate warmup can be used to push the optimization trajectory towards a flatter region of parameter space. \newline {\bf Bottom Row:} Full time evolution of curvature and learning rate for select runs (labeled with letters in the top row). {\bf A:} The non-BN variant diverges within 3 steps when trained at learning rate $.1$. {\bf B:} Same configuration as (A), but when warmup is used the curvature gradually decreases during training allowing successful training with peak learning rate of $.1$. {\bf C:} MetaInit allows training without warmup at learning rate $.1$ by starting off optimization in a flatter region of parameter space. Despite starting out with very flat curvature, we observe progressive sharpening until the curvature fluctuates around the bound $2 / \eta$. {\bf D:} An example of a model diverging despite starting out in the ``stable'' region. The curvature grows early and eventually diverges before step 100. {\bf E:} An example of progressive sharpening when training with a small learning rate. {\bf F:} Learning rate warmup recovers from an even poorer initialization (the same point diverges without warmup). 
     }
     \label{fig:curvature_curves}
\end{figure}

\section{The Interaction between Learning Rate Warmup, Initialization and Curvature}
\label{sec:warmup}

 The success of learning rate warmup is inconsistent with conventional optimization wisdom, which traditionally suggests adapting the step size to the curvature (see for example the discussion around equation 2.4 in \cite{mccandlish2018empirical}). However, with the understanding that $\lambda_1$ is a dynamic quantity whose evolution is tightly coupled with the learning rate schedule, the benefits of a warmup period are more easily understood. We argue that the success of learning rate warmup follows naturally from two properties of training deep models: 
\begin{enumerate}
    \item Models diverge when the learning rate is too large relative to the $2 / \lambda_1$ bound.
    \item When the learning rate only slightly exceeds $2 / \lambda_1$ optimization is unstable until the parameters move to a region with smaller $\lambda_1$ \cite{wu2018sgd, lewkowycz2020large}.
\end{enumerate}

\begin{figure}
     \centering
     \includegraphics[width=\textwidth]{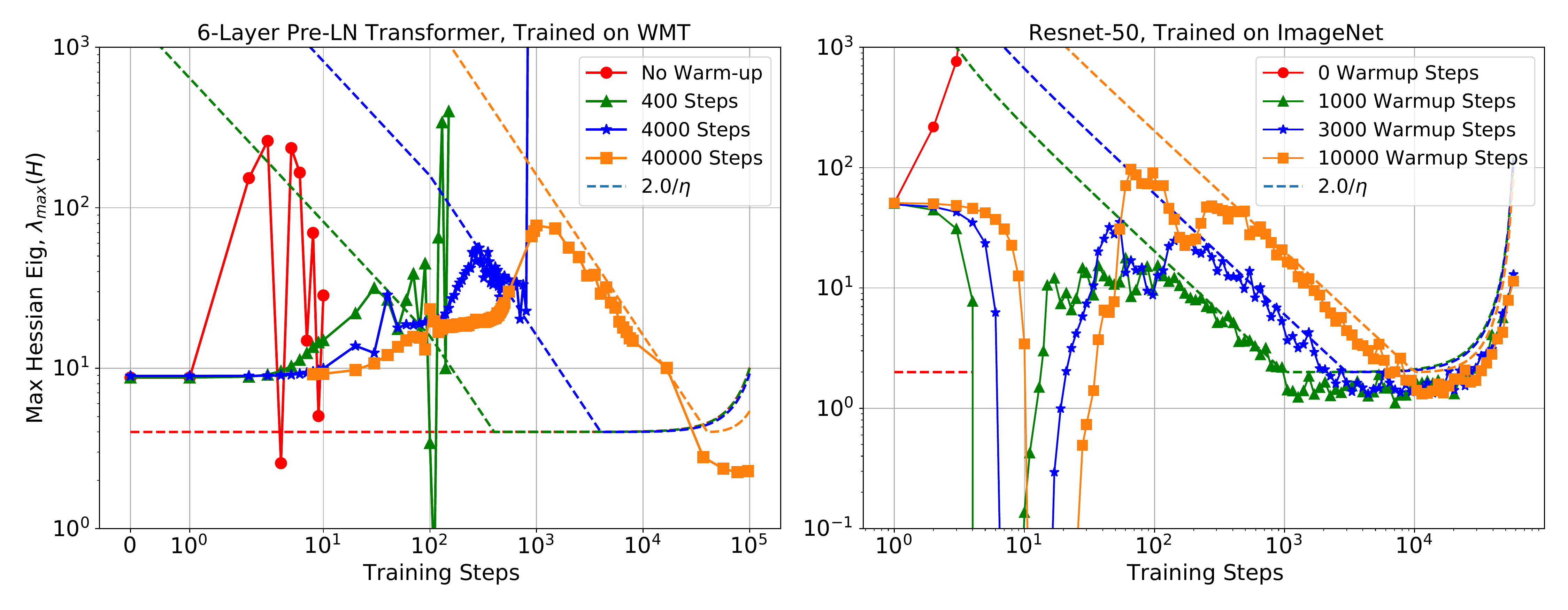}
     \caption{{\bf Learning rate warmup ``pushes'' the optimization trajectory towards regions of flatter curvature.} Solid lines correspond to the maximum eigenvalue of the Hessian throughout the training. Dashed lines correspond to $2/ \eta$. During the warmup period, $\lambda_1$ fluctuates close to the $2 / \eta$ bound. The WMT models diverge for 0, 400 and 4000 warmup steps. The pre-activation Resnet-50 model diverges quickly without warmup.  \label{fig:wmt-adaption}}
\end{figure}

The first criteria implies that we can't start $\eta$ off at too large of a value relative to $\lambda_1$ at initialization. The second criteria implies that gradually increasing $\eta$ can gradually ``push'' the parameters to a region of parameter space where optimization is stable (with lower values of $\lambda_1$). In Figure~\ref{fig:wmt-adaption} there is clear evidence for this ``pushing'', as during the warmup period the we see that $\lambda_1 \approx 2.0 / \eta$ holds for a large part  of the warmup phase. Furthermore, this approximation holds even as we vary the length of the warmup period. Other examples can be seen in Figure~\ref{fig:curvature_curves} (B and F), and Figure~\ref{fig:wrn_warmup_steps} in the appendix.

Warmup is not the only method capable of reducing $\lambda_1$ during training, one can instead initialize the model in a region where $\lambda_1$ starts off small. Consider for example, the points A, B and C in Figure~\ref{fig:curvature_curves}. Each point shows optimization of a non-BN WideResnet with peak learning rate of $.1$. In (A) we see the model diverges within 3 steps without warmup using the default initialization. In (B) we see that a linear warmup period results in $\lambda_1$ progressively decreasing until the peak step size of $.1$ is reached at step 1000, with no divergence occurring. Finally in (C) we initialize the same model with MetaInit, at which point $\lambda_1$ is small at initialization, and the model can be trained at $\eta = .1$ without warmup.

\begin{figure}
     \centering
     \includegraphics[width=.95\textwidth]{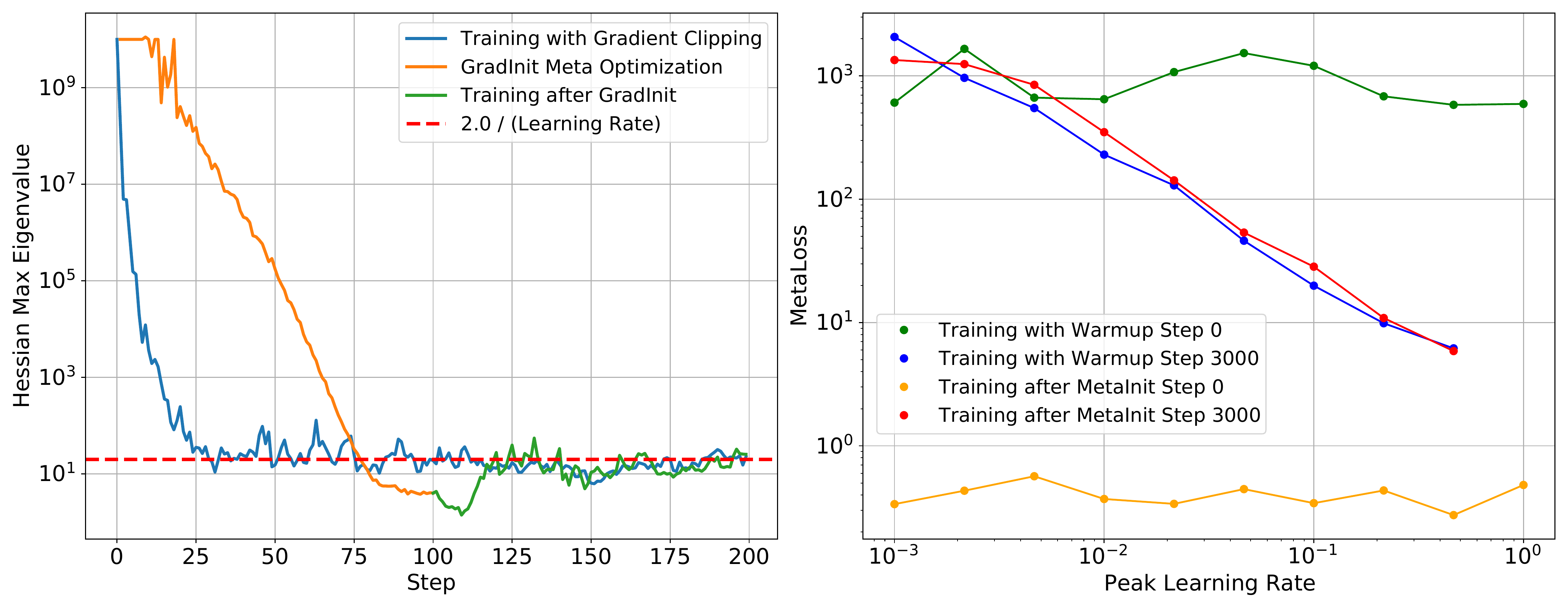}
     \caption{{\bf The mid-training conditioning is determined by the learning rate, not on the initialization method used. } {\em (a):} We plot the Hessian max eigenvalue during training for two models, a DenseNet model trained with learning rate warmup and gradient clipping, and the same model initialized with GradInit. We also plot the Hessian max eigenvalue during the GradInit meta optimization process. Both GradInit and training with  warmup are able to reduce the large curvature of the model. The GradInit algorithm initializes SGD in a flatter region, however this effect only last for a few steps of SGD, at which point the curvature of both the GradInit model and the model trained with warmup are nearly identical. {\em (b):} We plot the MetaInit MetaLoss during training for two groups of models, ones initialized with MetaInit and ones trained with learning rate warmup. Although the two groups start training with dramatically different values of the MetaLoss, after 3000 steps of SGD the MetaLoss of the two groups of models is almost completely determined by the learning rate used in training, not on the model initialization.\label{fig:gradinit_gradclip_comp}}
\end{figure}

Similar to the aforementioned MetaInit, the success of related initialization strategies can be explained by reduced $\lambda_1$ early in training. In Figure~\ref{fig:gradinit_gradclip_comp} (left) we look at the evolution of $\lambda_1$ during the GradInit meta optimization process and compare this with simply training the same model using gradient clipping\footnote{Similar to warmup, gradient clipping reduces the step size in regions of large curvature.}. Both methods result in $\lambda_1$ decreasing dramatically, after which $\lambda_1$ hovers around $2 / \eta$. Notably, GradInit starts regular training off at $\lambda_1$ significantly below the $2 / \eta$ bound, however the curvature quickly increases within a few steps. Given that initialization and warmup serve similar roles in reducing $\lambda_1$, we expect to be able to achieve similar performance using the two methods. As shown in Tables~\ref{tab:densenet} and ~\ref{tab:fixup} we can easily match key advances in this field by applying learning rate warmup alone (see Appendix for experimental details).

Beyond controlling $\lambda_1$ mid-training, the learning rate $\eta$ controls more general conditioning measures of the loss surface. For example, in Figure~\ref{fig:gradinit_gradclip_comp} we observe that even the MetaInit {\em gradient quotient}---the conditioning measure directly optimized by this initialization strategy---is controlled by $\eta$ mid training. This again provides further evidence that the primary benefit of this initialization method is to reduce $\lambda_1$ at initialization. As shown, any gains by optimizing the more general gradient quotient must be short lived as the initialization has no control over the long term value.

\begin{table}[]
\begin{center}
\begin{tabular}{lllll}
\hline
Model        & Method           & $Acc_0$   & $Acc_{Best}$ &  \\
\hline
DenseNet-100 & Kaiming-Clip 1.0 & 35.43 & 93.97   &  \\
DenseNet-100 & GradInit         & 37.19 & \bf{94.85}   &  \\
DenseNet-100 & Kaiming-Clip 6.0 & \bf{39.0}  & 94.65   &  \\
\hline
\end{tabular}
\caption{A comparison between the GradInit method and Kaiming initialization with gradient clipping on the DenseNet-100 architecture. The Kaiming-Clip 1.0 and GradInit rows were taken directly from \citet{zhu2021gradinit}, while the Kaiming-Clip 6.0 row is generated by running the open sourced Kaiming-Clip 1.0 baseline but changing the gradient clip coefficient from 1.0 to 6.0. The value of 6.0 was picked after we noticed that the gradient norms mid training were concentrated around 5.0, so the effect of setting the clip coefficient to 1.0 was artificially lowering the learning rate by a factor of 1/5 relative to the GradInit run. When this clip coefficient is changed to 6.0 the gap between the two methods shrinks significantly.\label{tab:densenet}}
\end{center}
\end{table}

\begin{table}[]
\begin{center}
\begin{tabular}{lllll}
\hline
Model  & Dataset  & Method   & $Acc$ &  \\
\hline
WideResnet 28-10 (w/o BN) & CIFAR-10 & Warmup 1000 & 97.2  &  \\
WideResnet 28-10 (w/o BN) & CIFAR-10 & MetaInit & 97.1  &  \\
\hline
Resnet-50 (w/o BN) & ImageNet & Fixup & 76.0 \\
Resnet-50 (w/o BN) & ImageNet & MetaInit & 76.0 \\
Resnet-50 (w/o BN) & ImageNet & GradInit & 76.2 \\
Resnet-50 (w/o BN) & ImageNet & Warmup 1000 & 76.2 \\
\hline
Transformer 6L (w/o LN) & WMT & Warmup + Clip + .25x Init & 27.10 (BLEU) \\
Transformer 6L (w/ LN) & WMT & LayerNorm & 27.01 (BLEU) \\
\hline

\end{tabular}
\caption{{\bf Learning rate warmup can match the performance of recent advances in initialization research.} The non-warmup results were all taken from their respective original works, MetaInit from \citet{dauphin2019metainit}, GradInit from \citet{zhu2021gradinit} and Fixup from \citet{zhang2019fixup}. We found in all cases that warmup and gradient clipping could be leveraged to recover from bad initializations, offering similar benefits to these better initialization strategies. For the 6-layer Transformer w/o LayerNorm we found the default initialization to exhibit extreme curvature, with initial gradient norm on the order of $10^{20}$. We found that warmup was insufficient in this case for any training to occur at this point, however scaling the default initialization down by a factor 4 was enough to recover the performance of the more sophisticated Fixup initialization.\label{tab:fixup}}
\end{center}
\end{table}

\section{The Effects of Curvature on Batch Size Scaling} \label{sec:bs_scaling}

 So far, we discussed how large loss curvature limits the range of stable learning rates for training. In this section, we highlight how these limits on usable learning rates affect the model's ability to effectively leverage larger batch sizes. Previous research has studied the interplay of the loss curvature and batch size scaling from various different perspectives. Most notably, \citet{shallue2018measuring} observe that increasing the batch size yields consistent improvements in training speed until a (problem-dependent) critical batch size is reached; increasing the batch size beyond this threshold yields diminishing improvements in training speed. \citet{zhang2019algorithmic} observe that a simple Noisy Quadratic Model (NQM) is able to capture the empirical behavior observed in \cite{shallue2018measuring}. Similarly, \citet{mccandlish2018empirical} use quadratic approximations to the loss to provide a closed form expression for the critical batch size as a function of the loss Hessian and the covariance of the stochastic gradient. We contribute to this literature by highlighting the role of $\lambda_1$ in the batch size scaling behavior of the model.

For this analysis, we focus on three of the WideResnet variants considered in Figure~\ref{fig:main_plot}---the BatchNorm model (a low curvature model), the non BatchNorm model (with moderate curvature), and the non BatchNorm model with 1.5X init scaling (with high curvature). We train these models while sweeping both the learning rate and the batch size. \footnote{We sweep for the optimal learning rate on a log-scale grid between $10^{-3}$ and $1$. For batch size, we sweep over powers of 2 from $16$ to $4096$.} We then measure the number of training steps required to reach $85\%$ validation accuracy, and the optimal learning rate found for each batch size. Similar to \cite{shallue2018measuring}, we normalize the plotted steps to $85\%$ accuracy by the value measured at batch size $64$. 

\begin{figure}[h]
     \centering
     \includegraphics[width=\textwidth]{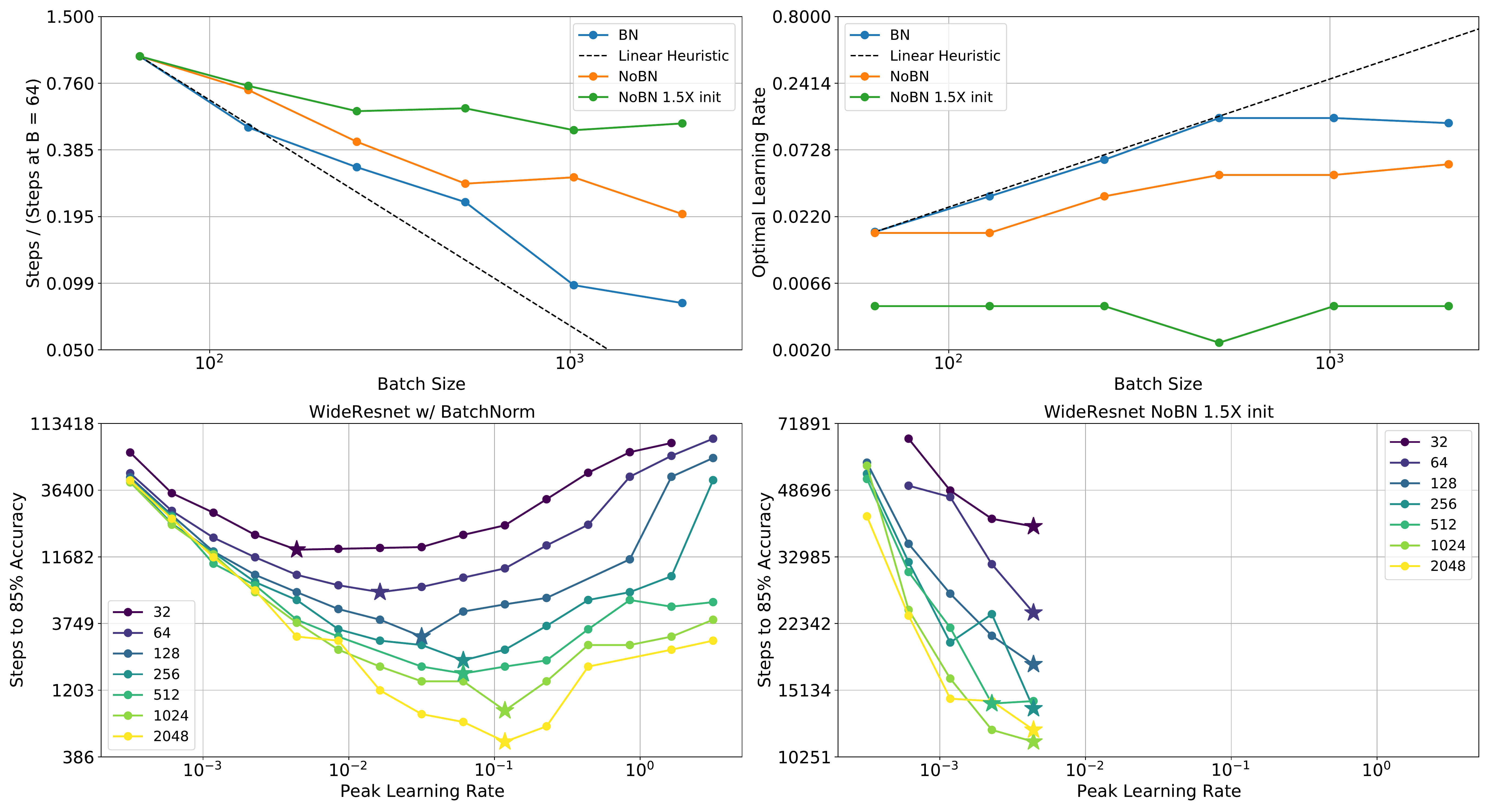}
     \caption{{\bf Large curvature models scale poorly with batch size.} The four plots explore how the WideResnet models from Figure~\ref{fig:main_plot} scale with batch size. We look at three models, a low curvature model (WideResnet with BatchNorm) a medium curvature model (WideResnet without BatchNorm), and a high curvature model (WideResnet w/o BatchNorm and 1.5 init scaling). The low curvature model exhibits almost perfect (linear) scaling as the batch size increases, with the optimal learning rate increasing almost linearly with the batch size. The high curvature model shows almost no speedups at larger batch sizes, with the optimal learning rate fixed at the largest value with does not diverge. {\em Top left:} Steps required for each model to reach 85\% accuracy, normalized by the steps required at batch size 64. {\em Top Right:} Optimal learning rate for each batch size. {\em Bottom Left:} Steps to 85\% accuracy for each learning rate, broken down by the batch size for the BatchNorm model. {\em Bottom Right:} Steps to 85\% accuracy for the non BatchNorm model with 1.5X init scaling. }
       \label{fig:wrn_bs}
\end{figure}

The results are shown in Figure~\ref{fig:wrn_bs}. A few observations are in order: The low curvature model shows almost linear speedups in training speed as the batch size increases. In contrast, the high curvature model exhibits only minimal improvements in training speed with larger batch sizes. These scaling differences are closely mirrored by how the optimal learning rate $\eta^*$ changes with the batch size: for the low curvature model $\eta^*$ increases linearly with the batch size, while for the high curvature model $\eta^*$ is fixed around $3 \times 10^{-3}$. Notably, for the high curvature model $\eta^*$ is almost always the largest non-divergent value---a clear indication that the high loss curvature slows down training by preventing larger values from being used.

A clear picture emerges from these observations. Previous research suggests that in order to effectively leverage larger batch sizes, one has to increase the learning rate in tandem with the batch size \cite{jastrzkebski2017three, goyal2017accurate, shallue2018measuring, mccandlish2018empirical}. Our results suggest that large values of $\lambda_1$ place a sharp limit on the maximum the learning rate possible and therefore, limit the model's ability to leverage data parallelism effectively. For the high curvature model the optimal learning rate $\eta^*$ is always close to the largest non-divergent learning rate---we believe this to be a useful diagnostic to detect if the training speed of a model is limited by $\lambda_1$ being too large. In such scenario, our results from Section \ref{sec:warmup} suggest that learning rate warm-up or better initialization algorithms can be used to improve the model's scaling behavior with respect to the batch size.

\section{Limitations}
\label{sec:limitations}

Our analysis has focused primarily on models trained with SGD with momentum. This decision was motivated to reduce additional confounds that arise when using adaptive preconditioning. Notably, it is unclear what the analogue of $\lambda_1 \leq 2 /\eta$ should be for a model trained with Adam. In the appendix, we provide evidence that loss curvature adaption to the learning rate does occur even for Transformer models trained with Adam, and that learning rate warmup results in the similar effect of the optimization trajectory being ``pushed'' to flatter regions. However we leave a deeper analysis into this for future work.

Finally, while our experiments certainly raise questions about the efficacy better model initialization has on further accelerating training, our measurements has focused primarily on the (lack of) influence initialization has on $\lambda_1$ mid training. It is possible that better initializations could have lasting influence on the broader Hessian eigenspectrum (for example improving the ration $\lambda_k / \lambda_1$ for smaller eigenvalues $\lambda_k$) and that our analysis is missing such an effect.

\section{Conclusion}

Through extensive empirical experiments measuring the evolution of the loss sharpness during training, we have demonstrated how different methods such as initialization, learning rate warmup, and normalization all enable higher learning rates to be used (without causing divergence) by reducing $\lambda_1$ during training. It is noteworthy that two of the most popular models we investigated (the popular variants of the Resnet-50 and WideResnet 28-10) did not benefit from learning rate warmup, and exhibited small values of $\lambda_1$ throughout training at the learning rates we considered. Thus researchers and practitioners who primarily work with well-optimized architectures might never notice a benefit from using warmup. However, even seemingly trivial modifications to a working architecture can easily result in large values of $\lambda_1$ and thus instability early in training---a naive response to such a situation would be to dramatically reduce the learning rate or, even worse, abandon the modification being investigated all together. We hope the perspective presented in this work can help future researchers better navigate such situations, either through investigating different initializations, applying warmup and gradient clipping, or changing the location of normalization layers in the model.

Perhaps the most striking feature of our results is how consistently $\lambda_1$ adapts to the learning rate, both in cases where $\lambda_1 < 2.0 / \eta$ and where $\lambda_1 > 2.0 / \eta$. While we are not the first to observe this adaptation, we have provided large scale empirical evidence for it under far less restrictive settings than prior work. This adaption of $\lambda_1$ to the learning rate is in direct contradiction with conventional optimization wisdom, which would instead recommend choosing the step size based on measured values of $\lambda_1$. Moving forward, much more work is needed to better understand why this adaptation occurs and to further understand the implications it has for optimizer design, step size selection, and model initialization.

\section{Societal Impact}
\label{sec:soc_impact}
 
Our results provide new understanding on neural network optimization that have the potential to accelerate research into better models through a more principled understanding of model tuning. While normalization free networks have been proposed recently, our work analyzes these methods from the lens of loss curvature, and enhances confidence in the working or failure modes of these networks. Normalization free networks improve power consumption at inference time, when deployed in large scale systems. We also suggest leveraging learning rate warmup as an alternative to computationally expensive initialization strategies, which require their own hyper parameter tuning. This would simplify the budget constraints, thereby reducing the environmental impact. While our efforts can potentially result in computational savings, this can actually sometimes lead to an increase in overall usage \citep{york2006ecological}.

Our results required numerous experiments in order to reach useful conclusions, and these machine learning workloads can sometimes result in significant carbon emissions \citep{patterson2021carbon}. That said, the cloud computing resources we used run on carbon-free power,\footnote{\url{https://www.gstatic.com/gumdrop/sustainability/24-7-explainer.pdf}} (although are not yet 24/7 carbon-free and still use some offsets).

Additionally, our work contributes to an increased understanding of machine learning methods, which have the potential to be used for benevolent or harmful purposes depending on the use case. Given the fundamental nature of our results, we do not believe the most likely uses are harmful.

\newpage
\bibliographystyle{abbrvnat}
\bibliography{main}

\begin{thebibliography}{38}
\providecommand{\natexlab}[1]{#1}
\providecommand{\url}[1]{\texttt{#1}}
\expandafter\ifx\csname urlstyle\endcsname\relax
  \providecommand{\doi}[1]{doi: #1}\else
  \providecommand{\doi}{doi: \begingroup \urlstyle{rm}\Url}\fi

\bibitem[Brock et~al.(2021)Brock, De, and Smith]{brock2021characterizing}
A.~Brock, S.~De, and S.~L. Smith.
\newblock Characterizing signal propagation to close the performance gap in
  unnormalized resnets.
\newblock \emph{arXiv preprint arXiv:2101.08692}, 2021.

\bibitem[Chelba et~al.(2013)Chelba, Mikolov, Schuster, Ge, Brants, and
  Koehn]{lm1b}
C.~Chelba, T.~Mikolov, M.~Schuster, Q.~Ge, T.~Brants, and P.~Koehn.
\newblock One billion word benchmark for measuring progress in statistical
  language modeling.
\newblock \emph{INTERSPEECH}, 2013.

\bibitem[Choi et~al.(2019)Choi, Shallue, Nado, Lee, Maddison, and
  Dahl]{choi2019empirical}
D.~Choi, C.~J. Shallue, Z.~Nado, J.~Lee, C.~J. Maddison, and G.~E. Dahl.
\newblock On empirical comparisons of optimizers for deep learning.
\newblock \emph{arXiv preprint arXiv:1910.05446}, 2019.

\bibitem[Cohen et~al.(2021)Cohen, Kaur, Li, Kolter, and
  Talwalkar]{cohen2021gradient}
J.~M. Cohen, S.~Kaur, Y.~Li, J.~Z. Kolter, and A.~Talwalkar.
\newblock Gradient descent on neural networks typically occurs at the edge of
  stability.
\newblock \emph{arXiv preprint arXiv:2103.00065}, 2021.

\bibitem[Dauphin and Schoenholz(2019)]{dauphin2019metainit}
Y.~N. Dauphin and S.~Schoenholz.
\newblock Metainit: Initializing learning by learning to initialize.
\newblock \emph{Advances in Neural Information Processing Systems},
  32:\penalty0 12645--12657, 2019.

\bibitem[Ghorbani et~al.(2019)Ghorbani, Krishnan, and
  Xiao]{ghorbani2019investigation}
B.~Ghorbani, S.~Krishnan, and Y.~Xiao.
\newblock An investigation into neural net optimization via hessian eigenvalue
  density.
\newblock In \emph{International Conference on Machine Learning}, pages
  2232--2241. PMLR, 2019.

\bibitem[Goyal et~al.(2017)Goyal, Doll{\'a}r, Girshick, Noordhuis, Wesolowski,
  Kyrola, Tulloch, Jia, and He]{goyal2017accurate}
P.~Goyal, P.~Doll{\'a}r, R.~Girshick, P.~Noordhuis, L.~Wesolowski, A.~Kyrola,
  A.~Tulloch, Y.~Jia, and K.~He.
\newblock Accurate, large minibatch sgd: Training imagenet in 1 hour.
\newblock \emph{arXiv preprint arXiv:1706.02677}, 2017.

\bibitem[He et~al.(2016{\natexlab{a}})He, Zhang, Ren, and Sun]{he2016deep}
K.~He, X.~Zhang, S.~Ren, and J.~Sun.
\newblock Deep residual learning for image recognition.
\newblock In \emph{Proceedings of the IEEE conference on computer vision and
  pattern recognition}, pages 770--778, 2016{\natexlab{a}}.

\bibitem[He et~al.(2016{\natexlab{b}})He, Zhang, Ren, and Sun]{he2016identity}
K.~He, X.~Zhang, S.~Ren, and J.~Sun.
\newblock Identity mappings in deep residual networks.
\newblock In \emph{European conference on computer vision}, pages 630--645.
  Springer, 2016{\natexlab{b}}.

\bibitem[Huang et~al.(2017)Huang, Liu, Van Der~Maaten, and
  Weinberger]{huang2017densely}
G.~Huang, Z.~Liu, L.~Van Der~Maaten, and K.~Q. Weinberger.
\newblock Densely connected convolutional networks.
\newblock In \emph{Proceedings of the IEEE conference on computer vision and
  pattern recognition}, pages 4700--4708, 2017.

\bibitem[Ioffe and Szegedy(2015)]{ioffe2015batch}
S.~Ioffe and C.~Szegedy.
\newblock Batch normalization: Accelerating deep network training by reducing
  internal covariate shift.
\newblock In \emph{International conference on machine learning}, pages
  448--456. PMLR, 2015.

\bibitem[Jastrz{\k{e}}bski et~al.(2017)Jastrz{\k{e}}bski, Kenton, Arpit,
  Ballas, Fischer, Bengio, and Storkey]{jastrzkebski2017three}
S.~Jastrz{\k{e}}bski, Z.~Kenton, D.~Arpit, N.~Ballas, A.~Fischer, Y.~Bengio,
  and A.~Storkey.
\newblock Three factors influencing minima in sgd.
\newblock \emph{arXiv preprint arXiv:1711.04623}, 2017.

\bibitem[Jastrzebski et~al.(2020)Jastrzebski, Szymczak, Fort, Arpit, Tabor,
  Cho, and Geras]{jastrzebski2020break}
S.~Jastrzebski, M.~Szymczak, S.~Fort, D.~Arpit, J.~Tabor, K.~Cho, and K.~Geras.
\newblock The break-even point on optimization trajectories of deep neural
  networks.
\newblock \emph{arXiv preprint arXiv:2002.09572}, 2020.

\bibitem[Kingma and Ba(2014)]{kingma2014adam}
D.~P. Kingma and J.~Ba.
\newblock Adam: A method for stochastic optimization.
\newblock \emph{arXiv preprint arXiv:1412.6980}, 2014.

\bibitem[Krizhevsky(2009)]{cifarKrizhevsky09learningmultiple}
A.~Krizhevsky.
\newblock Learning multiple layers of features from tiny images.
\newblock Technical report, University of Toronto, 2009.

\bibitem[Lewkowycz et~al.(2020)Lewkowycz, Bahri, Dyer, Sohl-Dickstein, and
  Gur-Ari]{lewkowycz2020large}
A.~Lewkowycz, Y.~Bahri, E.~Dyer, J.~Sohl-Dickstein, and G.~Gur-Ari.
\newblock The large learning rate phase of deep learning: the catapult
  mechanism.
\newblock \emph{arXiv preprint arXiv:2003.02218}, 2020.

\bibitem[Li et~al.(2017)Li, Xu, Taylor, Studer, and
  Goldstein]{li2017visualizing}
H.~Li, Z.~Xu, G.~Taylor, C.~Studer, and T.~Goldstein.
\newblock Visualizing the loss landscape of neural nets.
\newblock \emph{arXiv preprint arXiv:1712.09913}, 2017.

\bibitem[Liu et~al.(2020)Liu, Liu, Gao, Chen, and Han]{liu2020understanding}
L.~Liu, X.~Liu, J.~Gao, W.~Chen, and J.~Han.
\newblock Understanding the difficulty of training transformers.
\newblock \emph{arXiv preprint arXiv:2004.08249}, 2020.

\bibitem[McCandlish et~al.(2018)McCandlish, Kaplan, Amodei, and
  Team]{mccandlish2018empirical}
S.~McCandlish, J.~Kaplan, D.~Amodei, and O.~D. Team.
\newblock An empirical model of large-batch training.
\newblock \emph{arXiv preprint arXiv:1812.06162}, 2018.

\bibitem[Nado et~al.(2021)Nado, Gilmer, Shallue, Anil, and
  Dahl]{NadoEtAl2021_largeBatchReality}
Z.~Nado, J.~Gilmer, C.~J. Shallue, R.~Anil, and G.~E. Dahl.
\newblock A large batch optimizer reality check: Traditional, generic
  optimizers suffice across batch sizes.
\newblock \emph{CoRR}, abs/2102.06356, 2021.
\newblock URL \url{https://arxiv.org/abs/2102.06356}.

\bibitem[Papyan(2018)]{papyan2018full}
V.~Papyan.
\newblock The full spectrum of deepnet hessians at scale: Dynamics with sgd
  training and sample size.
\newblock \emph{arXiv preprint arXiv:1811.07062}, 2018.

\bibitem[Pascanu et~al.(2013)Pascanu, Mikolov, and
  Bengio]{pascanu2013difficulty}
R.~Pascanu, T.~Mikolov, and Y.~Bengio.
\newblock On the difficulty of training recurrent neural networks.
\newblock In \emph{International conference on machine learning}, pages
  1310--1318. PMLR, 2013.

\bibitem[Patterson et~al.(2021)Patterson, Gonzalez, Le, Liang, Munguia,
  Rothchild, So, Texier, and Dean]{patterson2021carbon}
D.~Patterson, J.~Gonzalez, Q.~Le, C.~Liang, L.-M. Munguia, D.~Rothchild, D.~So,
  M.~Texier, and J.~Dean.
\newblock Carbon emissions and large neural network training.
\newblock \emph{arXiv preprint arXiv:2104.10350}, 2021.

\bibitem[Pearlmutter(1994)]{pearlmutter1994fast}
B.~A. Pearlmutter.
\newblock Fast exact multiplication by the hessian.
\newblock \emph{Neural computation}, 6\penalty0 (1):\penalty0 147--160, 1994.

\bibitem[Russakovsky et~al.(2015)Russakovsky, Deng, Su, Krause, Satheesh, Ma,
  Huang, Karpathy, Khosla, Bernstein, et~al.]{russakovsky2015imagenet}
O.~Russakovsky, J.~Deng, H.~Su, J.~Krause, S.~Satheesh, S.~Ma, Z.~Huang,
  A.~Karpathy, A.~Khosla, M.~Bernstein, et~al.
\newblock Imagenet large scale visual recognition challenge.
\newblock \emph{International journal of computer vision}, 115\penalty0
  (3):\penalty0 211--252, 2015.

\bibitem[Santurkar et~al.(2018)Santurkar, Tsipras, Ilyas, and
  Madry]{santurkar2018does}
S.~Santurkar, D.~Tsipras, A.~Ilyas, and A.~Madry.
\newblock How does batch normalization help optimization?
\newblock \emph{arXiv preprint arXiv:1805.11604}, 2018.

\bibitem[Shallue et~al.(2018)Shallue, Lee, Antognini, Sohl-Dickstein, Frostig,
  and Dahl]{shallue2018measuring}
C.~J. Shallue, J.~Lee, J.~Antognini, J.~Sohl-Dickstein, R.~Frostig, and G.~E.
  Dahl.
\newblock Measuring the effects of data parallelism on neural network training.
\newblock \emph{arXiv preprint arXiv:1811.03600}, 2018.

\bibitem[Sivaprasad et~al.(2020)Sivaprasad, Mai, Vogels, Jaggi, and
  Fleuret]{pmlr-v119-sivaprasad20a}
P.~T. Sivaprasad, F.~Mai, T.~Vogels, M.~Jaggi, and F.~Fleuret.
\newblock Optimizer benchmarking needs to account for hyperparameter tuning.
\newblock In H.~D. III and A.~Singh, editors, \emph{Proceedings of the 37th
  International Conference on Machine Learning}, volume 119 of
  \emph{Proceedings of Machine Learning Research}, pages 9036--9045. PMLR,
  13--18 Jul 2020.
\newblock URL \url{http://proceedings.mlr.press/v119/sivaprasad20a.html}.

\bibitem[Vaswani et~al.(2017)Vaswani, Shazeer, Parmar, Uszkoreit, Jones, Gomez,
  Kaiser, and Polosukhin]{vaswani_transformers}
A.~Vaswani, N.~Shazeer, N.~Parmar, J.~Uszkoreit, L.~Jones, A.~N. Gomez, L.~u.
  Kaiser, and I.~Polosukhin.
\newblock Attention is all you need.
\newblock In \emph{Advances in Neural Information Processing Systems},
  volume~30, 2017.
\newblock URL
  \url{https://proceedings.neurips.cc/paper/2017/file/3f5ee243547dee91fbd053c1c4a845aa-Paper.pdf}.

\bibitem[Wu et~al.(2018)Wu, Ma, and E]{wu2018sgd}
L.~Wu, C.~Ma, and W.~E.
\newblock How sgd selects the global minima in over-parameterized learning: A
  dynamical stability perspective.
\newblock In \emph{Proceedings of the 32nd International Conference on Neural
  Information Processing Systems}, pages 8289--8298, 2018.

\bibitem[Xiong et~al.(2020)Xiong, Yang, He, Zheng, Zheng, Xing, Zhang, Lan,
  Wang, and Liu]{layernormstransformer}
R.~Xiong, Y.~Yang, D.~He, K.~Zheng, S.~Zheng, C.~Xing, H.~Zhang, Y.~Lan,
  L.~Wang, and T.~Liu.
\newblock On layer normalization in the transformer architecture.
\newblock In \emph{Proceedings of the 37th International Conference on Machine
  Learning, {ICML} 2020, 13-18 July 2020, Virtual Event}, volume 119 of
  \emph{Proceedings of Machine Learning Research}, pages 10524--10533. {PMLR},
  2020.

\bibitem[Yang et~al.(2019)Yang, Pennington, Rao, Sohl-Dickstein, and
  Schoenholz]{yang2019mean}
G.~Yang, J.~Pennington, V.~Rao, J.~Sohl-Dickstein, and S.~S. Schoenholz.
\newblock A mean field theory of batch normalization.
\newblock \emph{arXiv preprint arXiv:1902.08129}, 2019.

\bibitem[York(2006)]{york2006ecological}
R.~York.
\newblock Ecological paradoxes: William stanley jevons and the paperless
  office.
\newblock \emph{Human Ecology Review}, pages 143--147, 2006.

\bibitem[Zagoruyko and Komodakis(2016)]{zagoruyko2016wide}
S.~Zagoruyko and N.~Komodakis.
\newblock Wide residual networks.
\newblock \emph{arXiv preprint arXiv:1605.07146}, 2016.

\bibitem[Zhang et~al.(2019{\natexlab{a}})Zhang, Li, Nado, Martens, Sachdeva,
  Dahl, Shallue, and Grosse]{zhang2019algorithmic}
G.~Zhang, L.~Li, Z.~Nado, J.~Martens, S.~Sachdeva, G.~Dahl, C.~Shallue, and
  R.~B. Grosse.
\newblock Which algorithmic choices matter at which batch sizes? insights from
  a noisy quadratic model.
\newblock \emph{Advances in neural information processing systems},
  32:\penalty0 8196--8207, 2019{\natexlab{a}}.

\bibitem[Zhang et~al.(2017)Zhang, Cisse, Dauphin, and
  Lopez-Paz]{zhang2017mixup}
H.~Zhang, M.~Cisse, Y.~N. Dauphin, and D.~Lopez-Paz.
\newblock mixup: Beyond empirical risk minimization.
\newblock \emph{arXiv preprint arXiv:1710.09412}, 2017.

\bibitem[Zhang et~al.(2019{\natexlab{b}})Zhang, Dauphin, and
  Ma]{zhang2019fixup}
H.~Zhang, Y.~N. Dauphin, and T.~Ma.
\newblock Fixup initialization: Residual learning without normalization.
\newblock \emph{arXiv preprint arXiv:1901.09321}, 2019{\natexlab{b}}.

\bibitem[Zhu et~al.(2021)Zhu, Ni, Xu, Kong, Huang, and
  Goldstein]{zhu2021gradinit}
C.~Zhu, R.~Ni, Z.~Xu, K.~Kong, W.~R. Huang, and T.~Goldstein.
\newblock Gradinit: Learning to initialize neural networks for stable and
  efficient training.
\newblock \emph{arXiv preprint arXiv:2102.08098}, 2021.

\end{thebibliography}

\newpage
\appendix
\section{Miscellaneous Figures}

\begin{figure}[h]
     \centering
    \includegraphics[width=\textwidth]{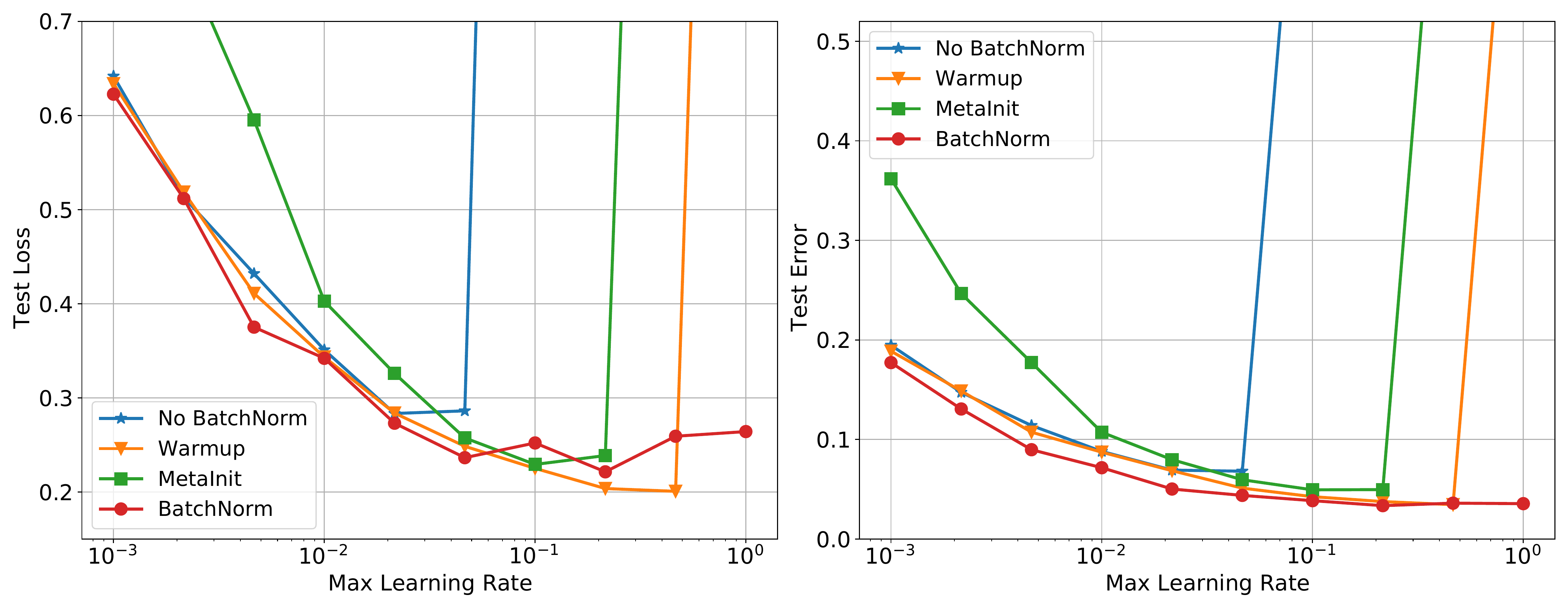}
     \caption{The test-time behavior of the different WideResnet models closely mirrors their training loss dynamics (Figure~\ref{fig:wideresnet_performance}, left).}
     \label{fig:wideresnet_test_performance}
\end{figure}

\begin{figure}
     \centering
     \includegraphics[width=\textwidth]{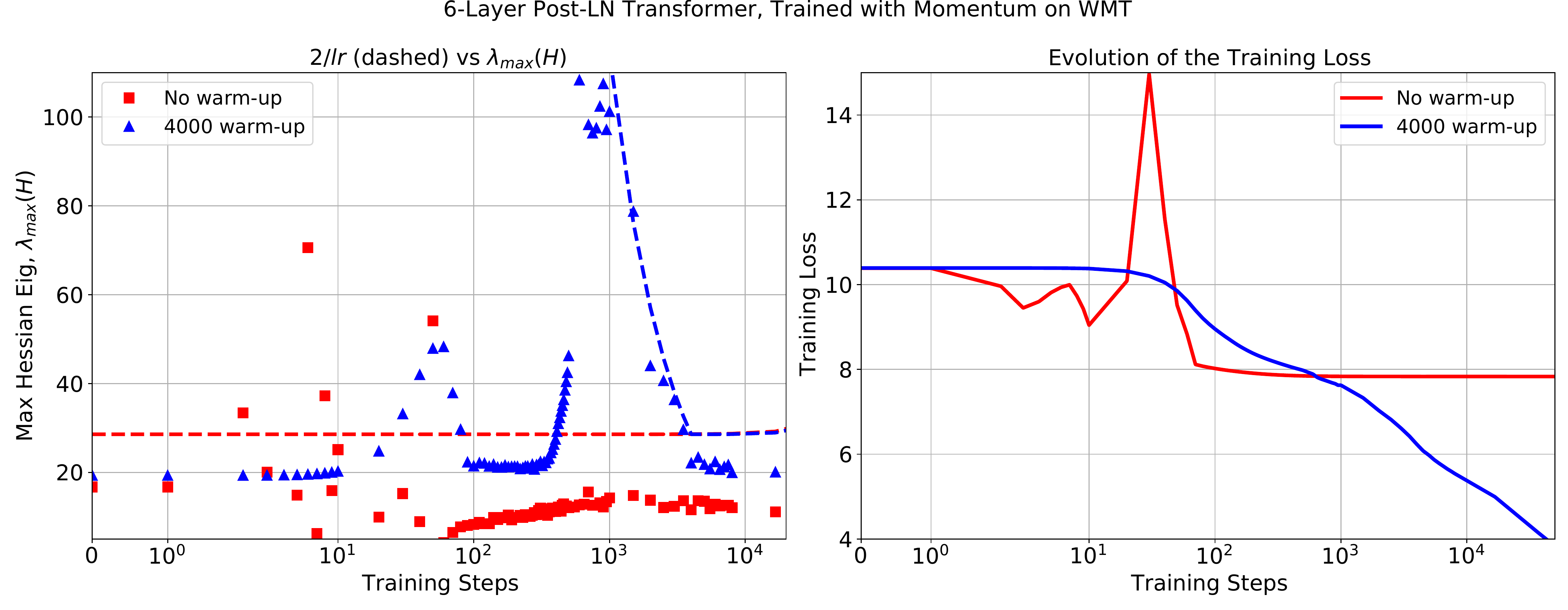}
     \caption{The curvature at initialization can be unreliable for predicting training stability. Left: Maximum eigenvalues of the Hessian throughout training. Dashed line correspond to $2/\eta$. Right: Evolution of the training loss for different models. The model with no warm-up (red) suffers from training instabilities even though it is below the stability threshold at initialization.\label{fig:unreliable_init}}
\end{figure}

\section{Performance of models in Figure 2}

\begin{figure}[h]
     \centering
     \begin{subfigure}[b]{0.49\textwidth}
         \centering
         \includegraphics[width=\textwidth]{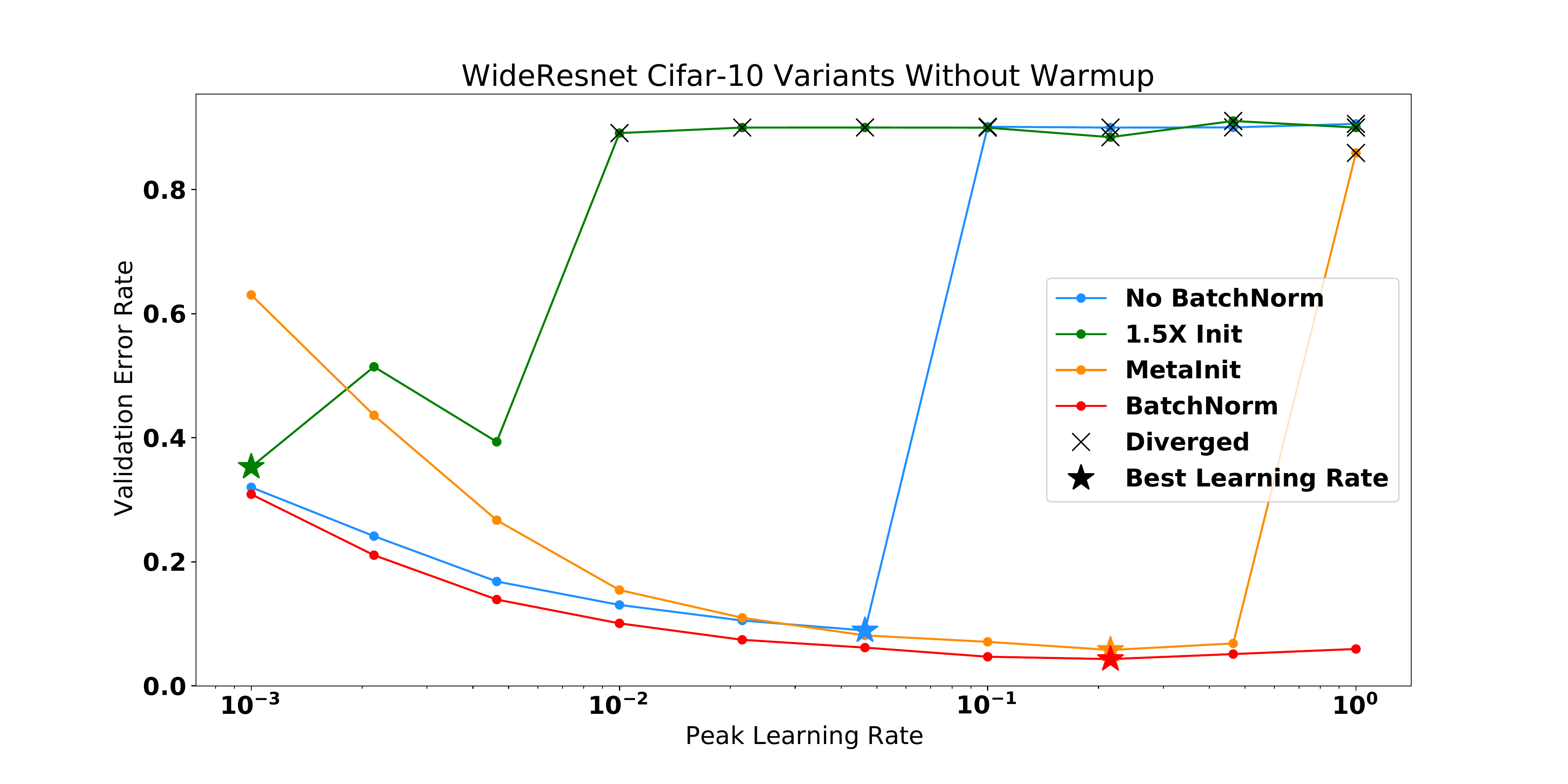}
         
     \end{subfigure}
     \hfill
     \begin{subfigure}[b]{0.49\textwidth}
         \centering
         \includegraphics[width=\textwidth]{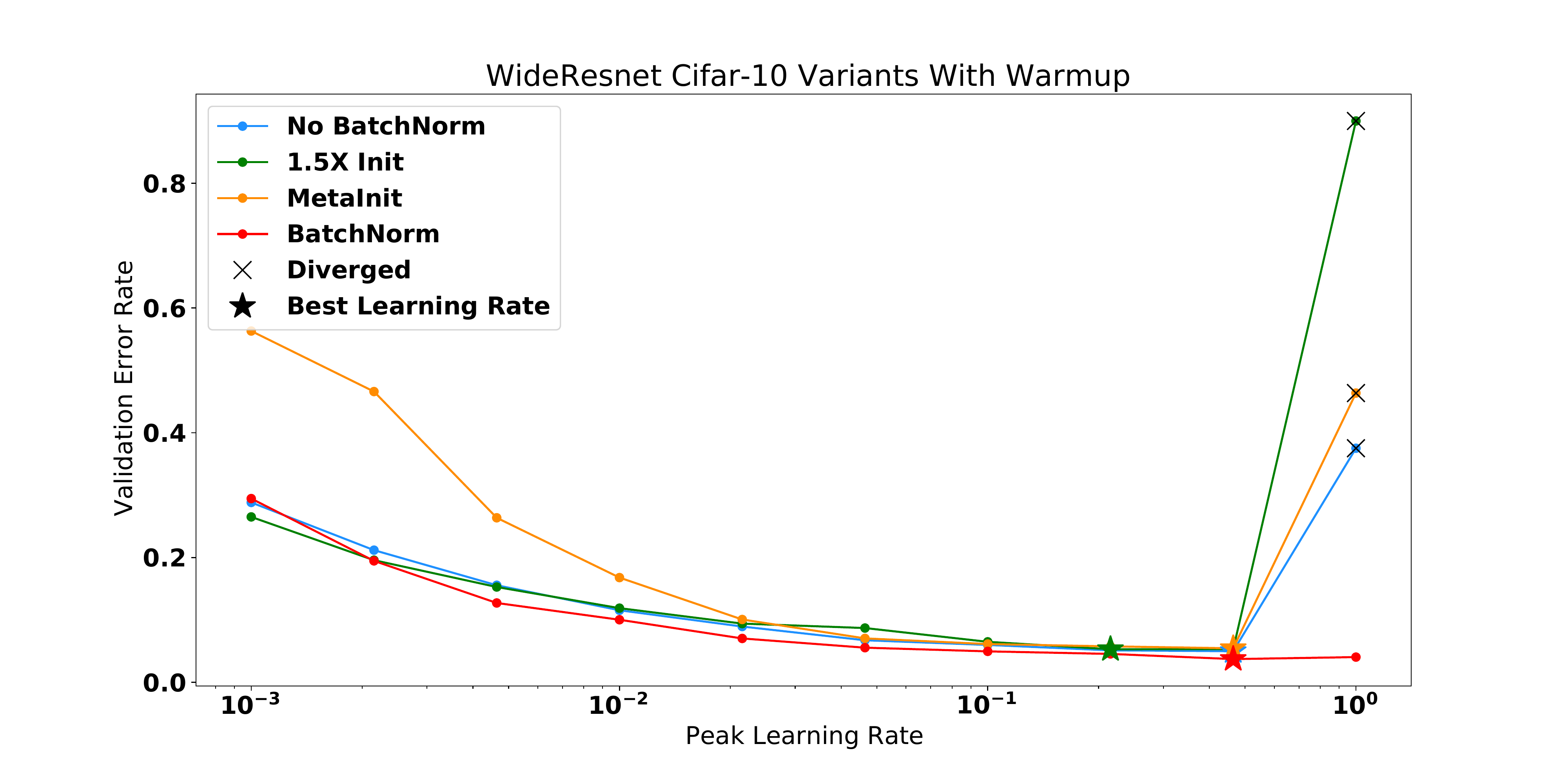}
     \end{subfigure}
     \caption{Performance of the WideResnet 28-10 models trained on Cifar10 shown in Figure 2 of the main text. (left) Performance of models trained without learning rate warmup. (right) Performance with learning rate warmup. When warmup is applied, all models reach comparable performance (though additional regularization is needed to match the generalization of the Batch Normalized models). \label{fig:wrn_train}}
\end{figure}

\begin{figure}[h]
     \centering
     \begin{subfigure}[b]{0.49\textwidth}
         \centering
         \includegraphics[width=\textwidth]{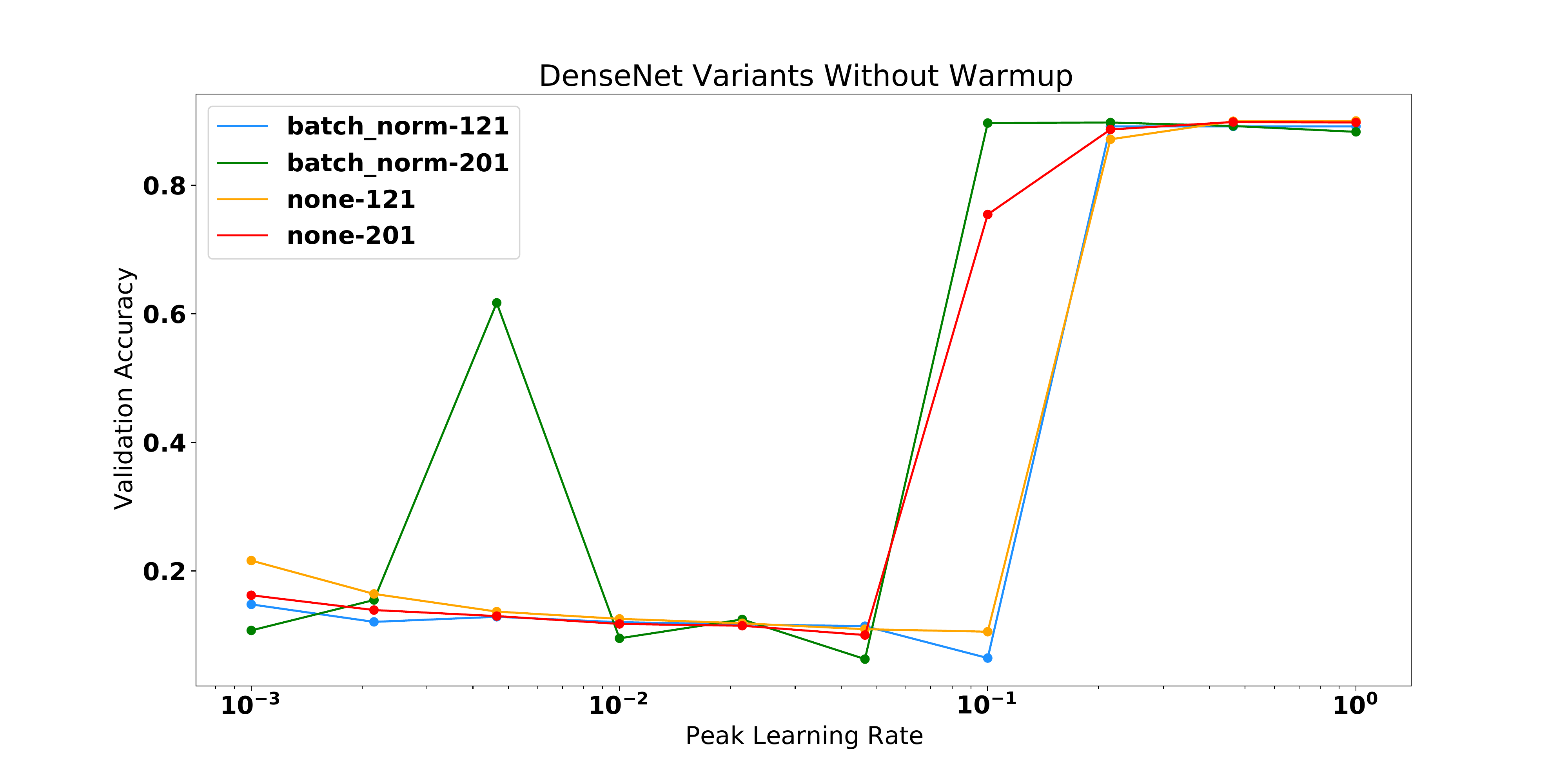}
         
     \end{subfigure}
     \hfill
     \begin{subfigure}[b]{0.49\textwidth}
         \centering
         \includegraphics[width=\textwidth]{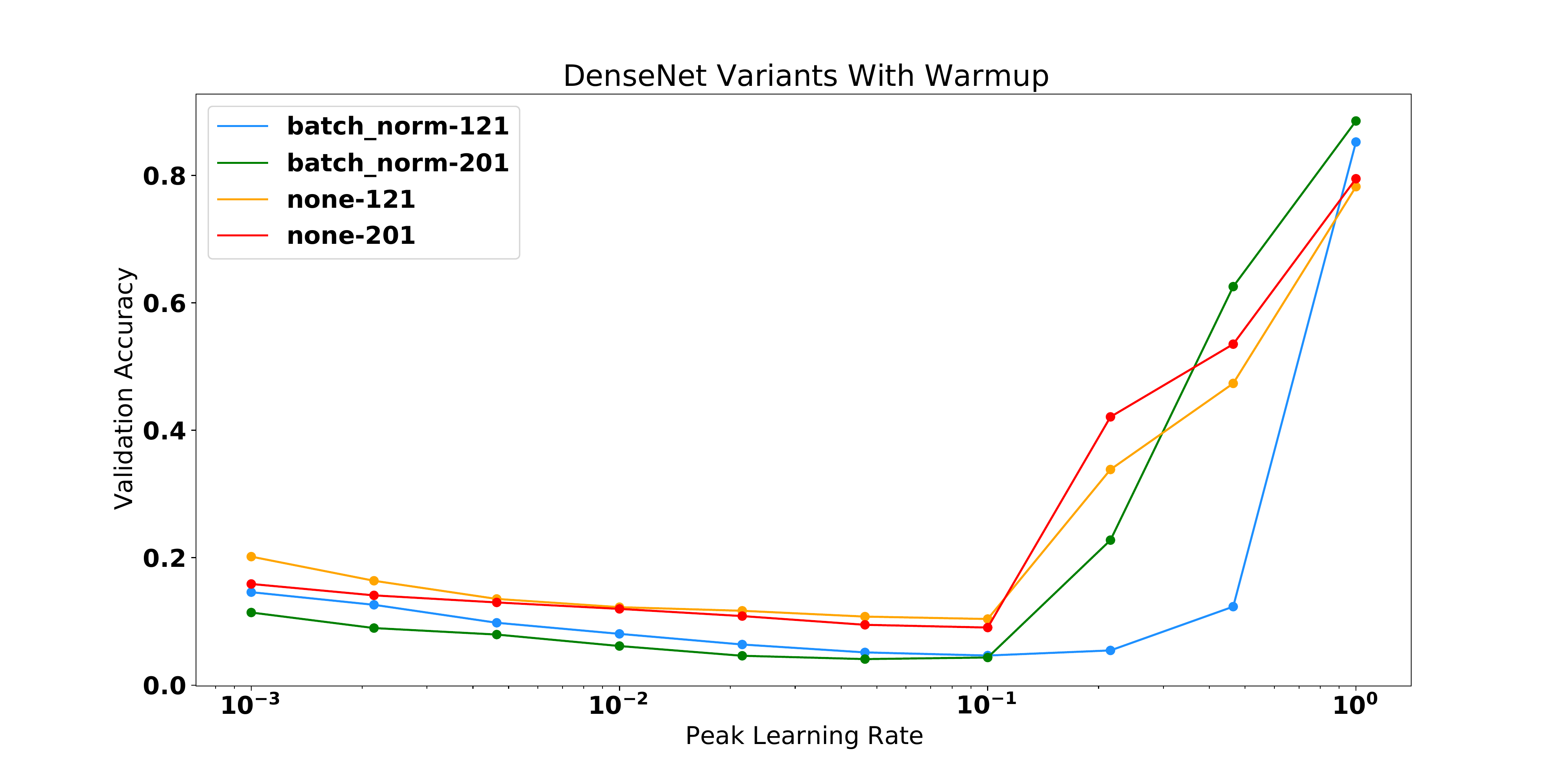}
     \end{subfigure}
     \caption{Performance of the DenseNet models trained on Cifar10 shown in Figure 2 of the main text. (left) Performance of models trained without learning rate warmup. (right) Performance with learning rate warmup.\label{fig:densenet_train}}
\end{figure}

\begin{figure}[h]
     \centering
     \begin{subfigure}[b]{0.49\textwidth}
         \centering
         \includegraphics[width=\textwidth]{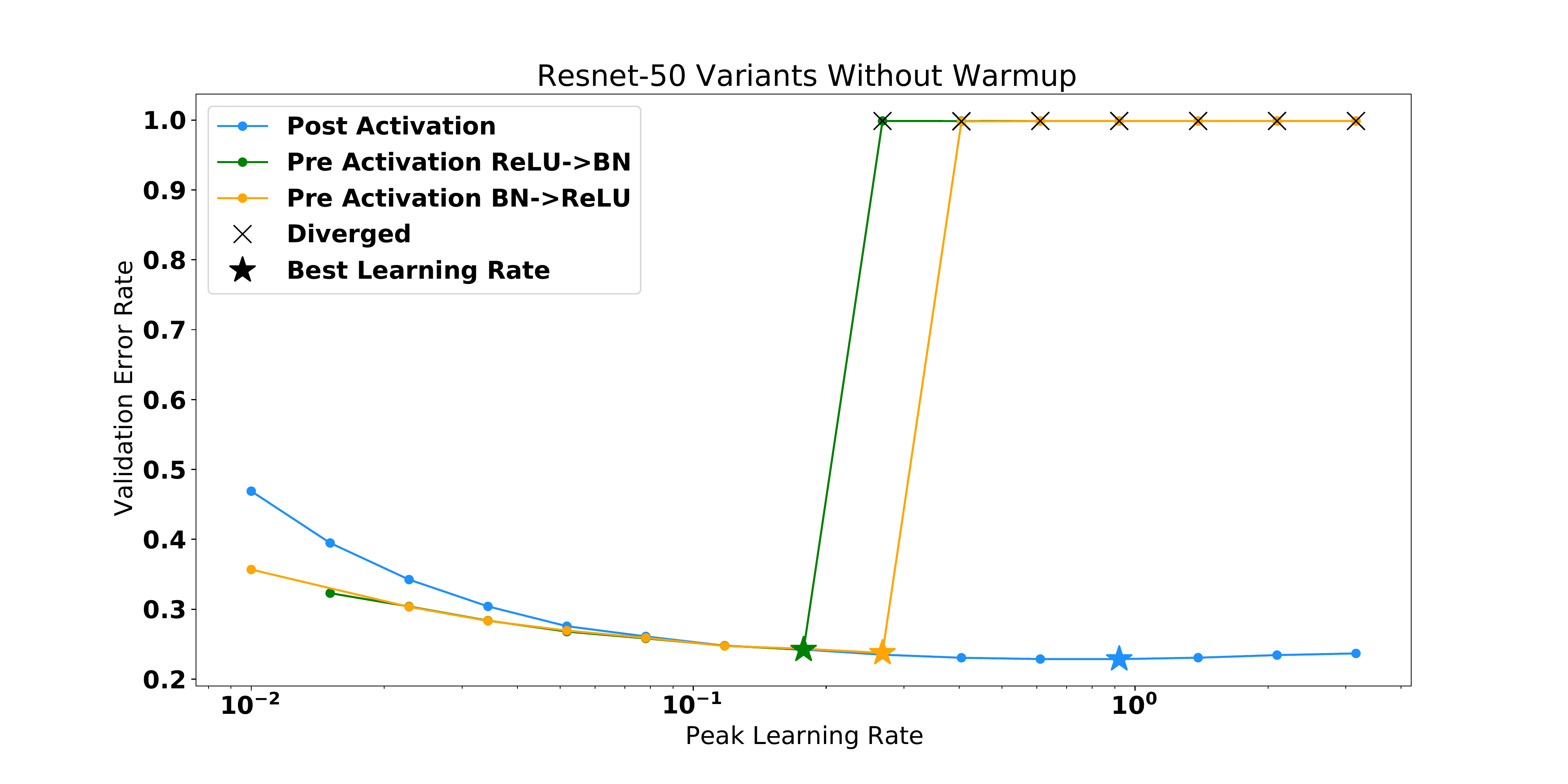}
         
     \end{subfigure}
     \hfill
     \begin{subfigure}[b]{0.49\textwidth}
         \centering
         \includegraphics[width=\textwidth]{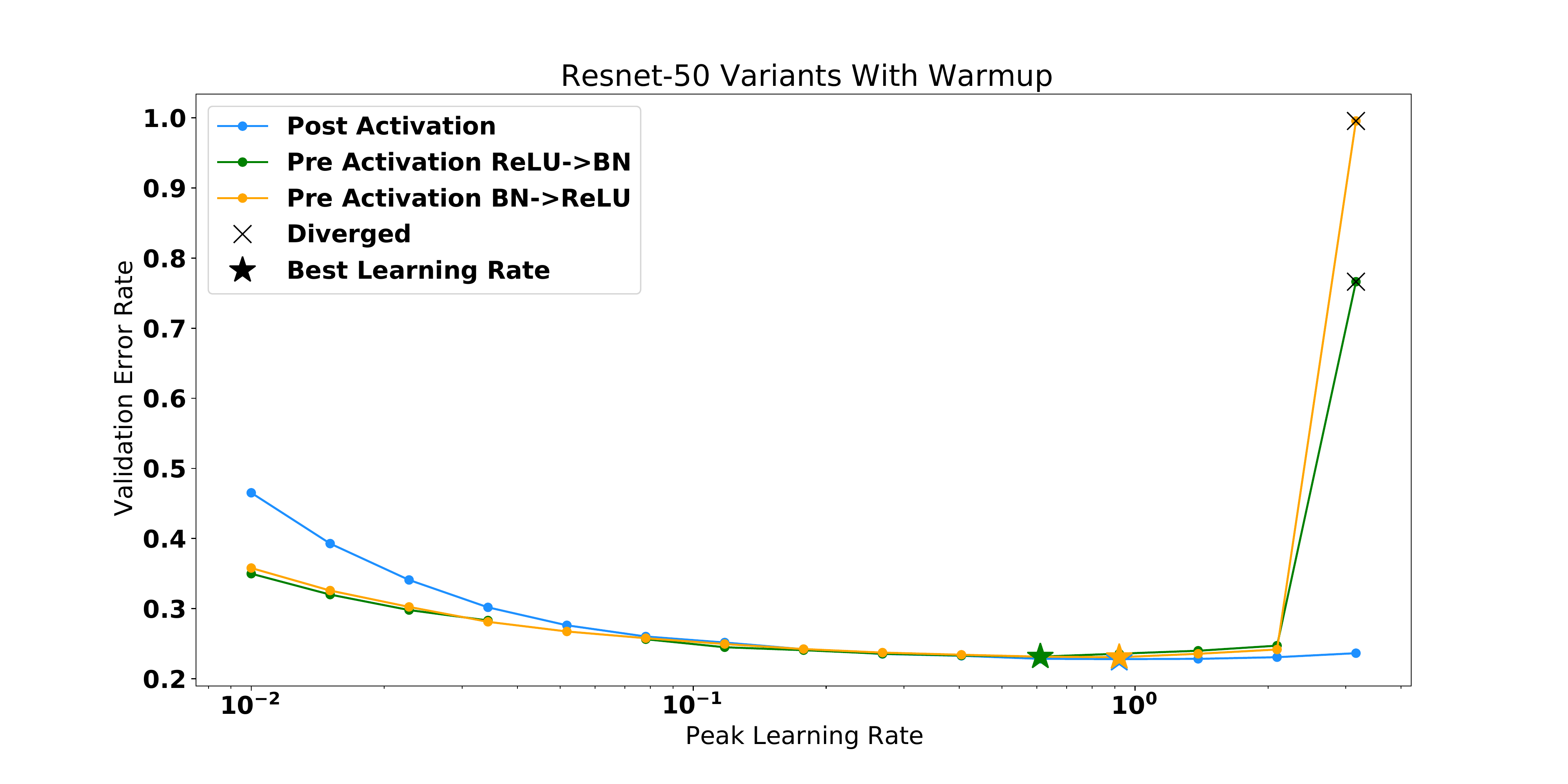}
     \end{subfigure}
     \caption{Performance of the Resnet50 models trained on Imagenet shown in Figure 2 of the main text. (left) Performance of models trained without learning rate warmup. (right) Performance with learning rate warmup.\label{fig:resnet_train}}
\end{figure}

\begin{figure}
     \centering
     \begin{subfigure}[b]{0.49\textwidth}
         \centering
         \includegraphics[width=\textwidth]{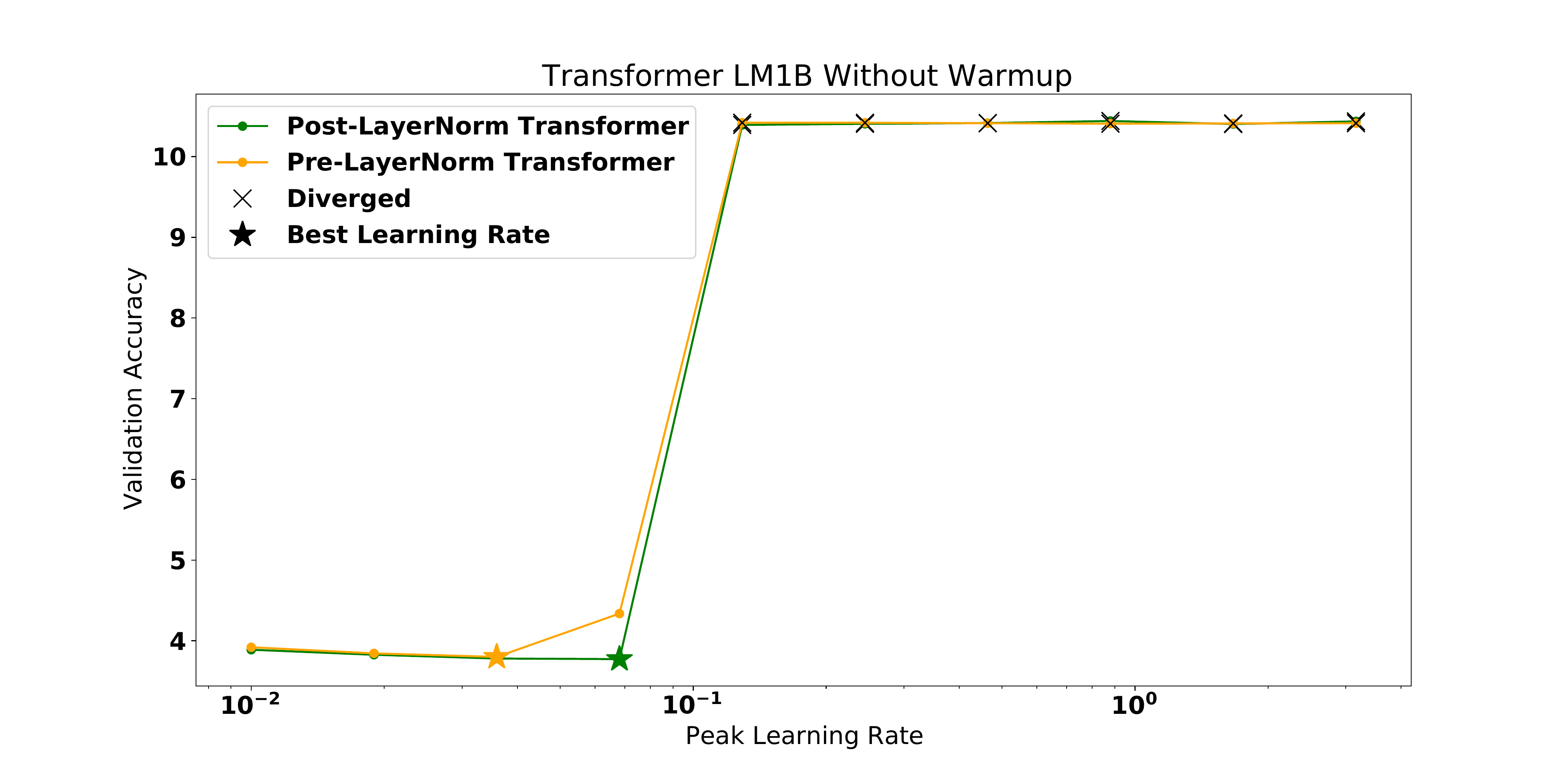}
         
     \end{subfigure}
     \hfill
     \begin{subfigure}[b]{0.49\textwidth}
         \centering
         \includegraphics[width=\textwidth]{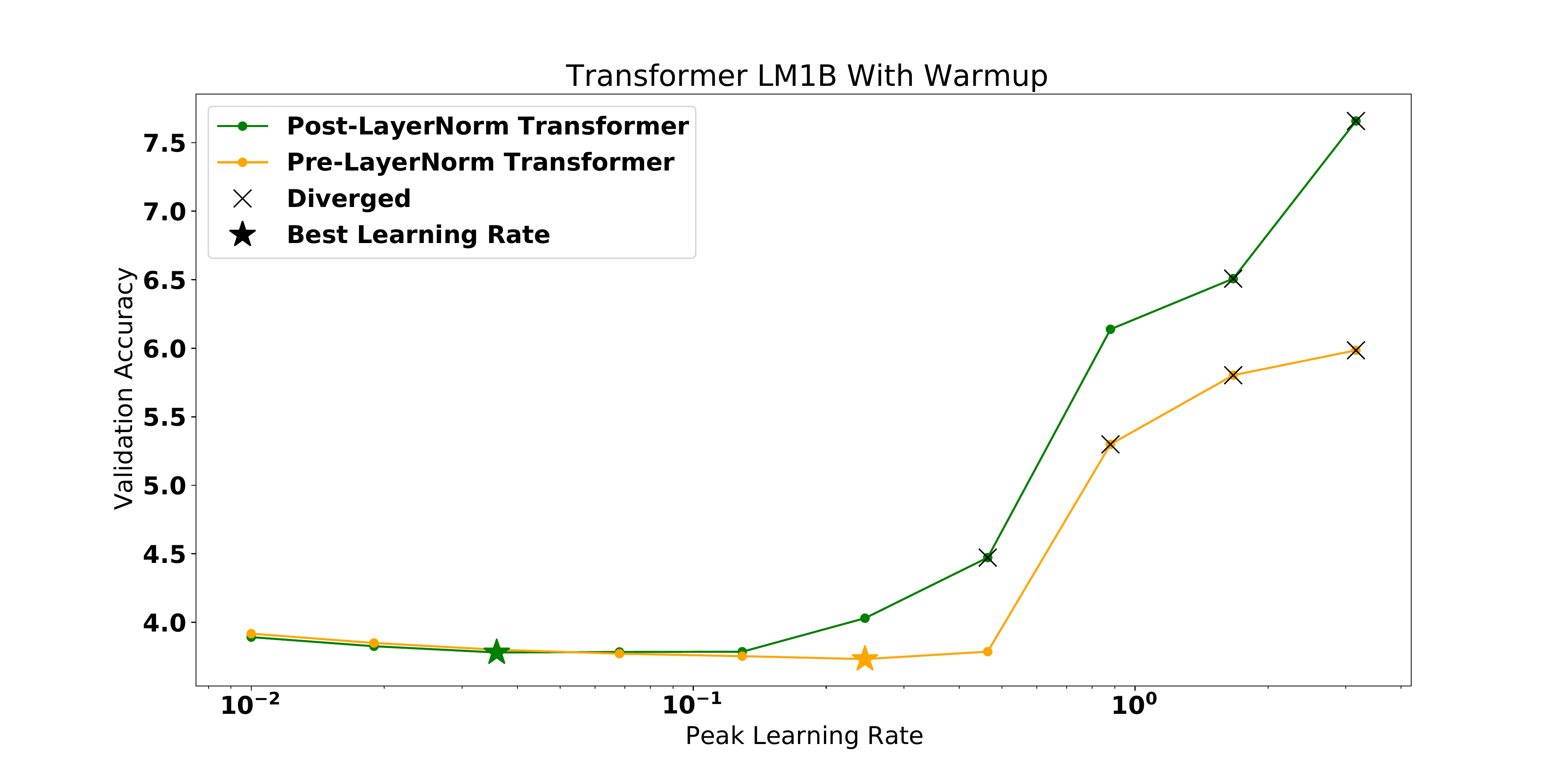}
     \end{subfigure}
     \caption{Performance of the Transformer models trained on LM1B shown in Figure 2 of the main text. (left) Performance of models trained without learning rate warmup. (right) Performance with learning rate warmup.\label{fig:transformer_train}}
\end{figure}

In this this section, we plot the performance vs learning rate for all of the models shown in Figure 2 of the main text. These are shown in Figures~\ref{fig:wrn_train}, \ref{fig:densenet_train}, \ref{fig:resnet_train}, and \ref{fig:transformer_train}. For models which diverged, we plot the best test performance achieved before divergence. In all settings, high curvature affects the final performance by limiting the use of higher learning rates.

\begin{figure}
     \centering
     \includegraphics[width=\textwidth]{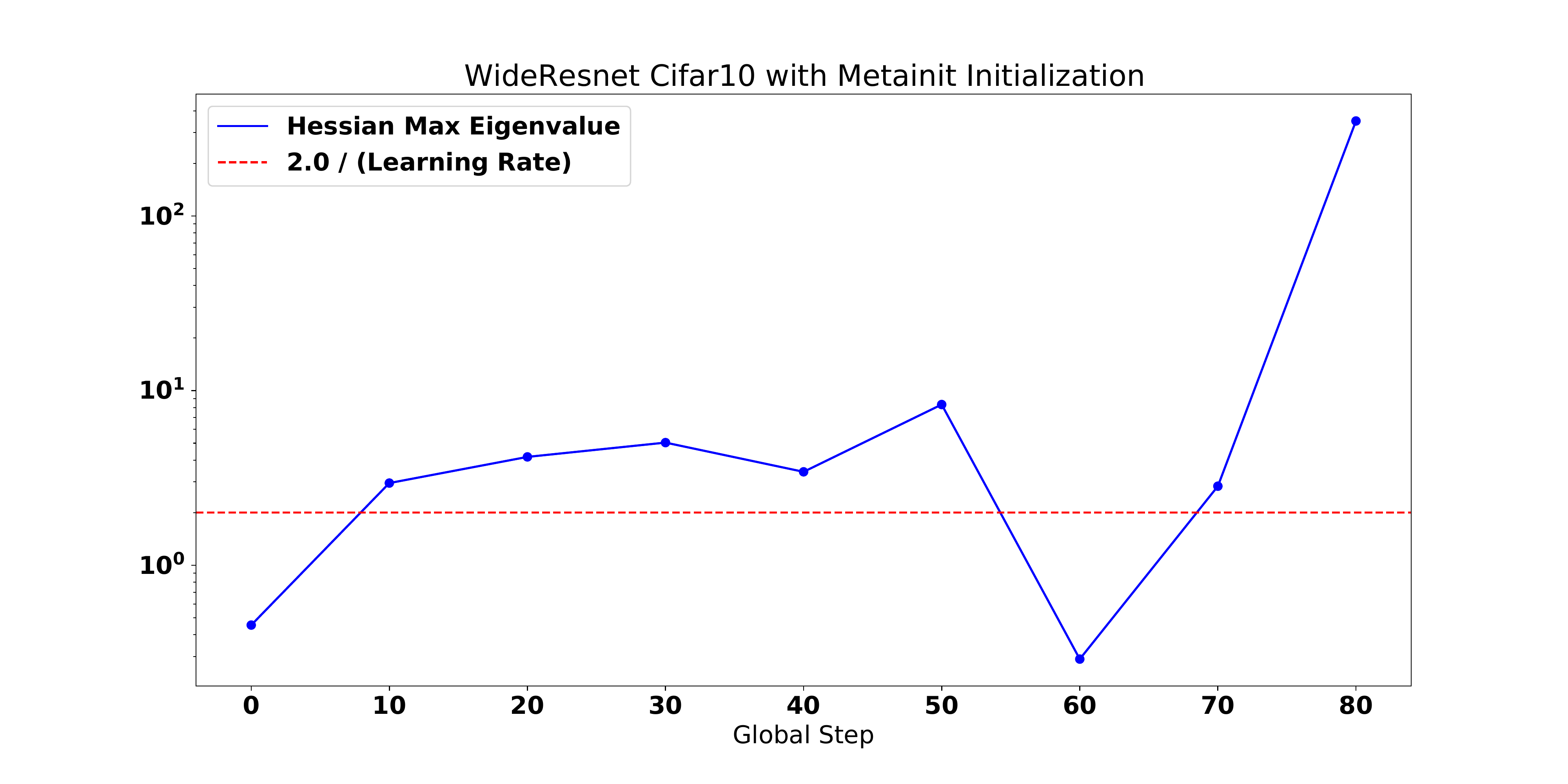}
     \caption{Training curve of the WideResnet 28-10 which diverged despite starting out in a region with $\lambda_1 < 2.0 / \eta$. The parameters quickly leave the stable region and end up in a region of higher curvature before diverging.\label{fig:metainit_divergence} }
\end{figure}

We also noted several models in Figure 2 which diverged despite training starting out in a stable region of parameter space. In Figure~\ref{fig:metainit_divergence} we plot the evolution of the loss sharpness during training, showing that it quickly enters a region where $\lambda_1 > 2.0 / \eta$ before diverging around step 90.

\section{Details on Computing the Hessian Eigenspectrum via Lanczos}

We use Lanczos iterations to estimate the top eigenvalue of the Hessian. Lanczos algorithm only requires Hessian-vector products which can be efficiently computed via Pearlmutter's trick \cite{pearlmutter1994fast}. Previous research has demonstrated that this approach provides a robust and scalable framework to examine the eigenvalues of the Hessian for large neural networks \cite{ghorbani2019investigation, papyan2018full}.  

For our WMT / LM1B experiments, we run the algorithm for $45$ steps while for image models, we use $40$ steps. When monitoring the evolution of the top eigenvalue as a function of the number of Lanczos steps, in all cases except one, we observe that the algorithm converges. For the case of Resnet with ReLU$\rightarrow$BN ordering, due to a very small eigengap between the top eigenvalue and the bulk, the convergence is significantly slower. We use $200$ Lanczos steps in this case to alleviate the issue.

It is well-known that Lanczos algorithm can suffer from numerical instabilities caused by finite-precision arithmetic. To alleviate these issues, Lanczos vectors are stored in float64 accuracy and we perform reorthogonalized at each step of the algorithm.

\section{Training Details for Tables 1 and 2}

\subsection{Neural Machine Translation}
Neural Machine Translation experiments are based on the Transformer models \citep{vaswani_transformers}. We use separate embeddings on encoder and decoder, and a common word piece vocabulary of size 32000. For depth, we use 6 layers on both encoder and decoder. For width, we experiment with two models, namely Transformer-Base and Transformer-Wide. For Transformer-Base, we use word embeddings with 512 dimensions, 8 heads and 2048 feed-forward dimension. For Transformer-Wide we use word embeddings with 1024 dimensions, 16 heads and 4096 feed-forward dimension. The experiments reported in Figures 3 and 5 use Transformer-Base. The experiments reported in  Table 2 use Transformer-Wide models trained with Adam \citep{kingma2014adam}. We sweep over warm-up, learning rate, gradient clipping and init\_scaling and optimize for validation loss to evaluate performance on test set BLEU reported in Table 2. All the models are trained for 60 epochs at batch size of 1024 for Transformer-Base models, and batch size of 512 for Transformer-Big models. We use dropout of 0.1, label smoothing of 0.1 and no weight decay for all these models. 

\subsection{DenseNets}

In Table 2 the ResNet-50 (w/o BN) architecture was trained for 100 epochs at batch size 512, with l2 regularization of 5e-5, dropout of .3. It was trained with SGD with nesterov momentum of .9 and learning rate of .2. We applied gradient clipping at global l2 norm of 5 and used linear learning rate warmup with warmup period of 1000 steps.

For Table 1, the DenseNet-100 model was trained using the Gradinit codebase \footnote{\url{https://github.com/zhuchen03/gradinit}} by modifying the supplied DenseNet script to apply gradient clipping of norm 6 and to use the default initialization instead of GradInit. 

\section{Training Details for Figure 2}

The WideResnet-28-10 models were trained with batch size of 1024 for 300 epochs. We applied the MixUp augmentation\citep{zhang2017mixup}. For learning rate warmup we used 1000 steps of linear warmup until the peak learning rate is achieved, at which point the learning rate is decayed according to the cosine schedule.

The DenseNet models were trained with batch size of 512 using the SGD optimizer with momentum of 0.9, weight decay of 5e-4, L2 regularization of 1e-4 and warmup of 1000 steps (for the models where warmup is used) followed by cosine decay. The models were trained for 200 epochs. For the DenseNet architecture we used growth\_rate of 32 and reduction of 0.5.   

The Resnet-50 models were trained with batch size of 2048 using the SGD optimizer with nesterov momentum of .9. The learning rate schedule was the same as in the WideResnet case, with linear warmup of 1000 steps followed by cosine decay. We applied label smoothing of .1 and used the standard flip plus crop for data augmentation.

The Transformer models on LM1B were trained at batch size 1024 using SGD with nesterov momentum of .9. We use embedding dimension of 512, 6 layers with 8 heads and MLP hidden dimension of 1024. The attention dropout rate was .1. The learning rate schedule followed the same recipe as in the Resnet cases.

\section{Curvature Adaptation with the Adam Optimizer}
The discussion in the main text focused primarily on models trained with SGD and momentum. In this appendix, we briefly examine if similar conclusions hold for optimizers such as Adam that use preconditioning. It is unclear a priori whether or not curvature adapation to the learning rate should occur for optimizers which apply preconditioning. However, given that Adam is a diagonal preconditioner applied to a non-diagonal Hessian, there may be some similar effects observed. 

Consider a simple quadratic loss 
\begin{align*}
L(\theta) = \frac{1}{2} \theta^T H \theta, \; H \succ 0. 
\end{align*}
where optimization is performed via preconditioned gradient descent with a fixed diagonal preconditioning matrix $D$:
\begin{align*}
    \theta_t &= \theta_{t - 1} - \eta D^{-1} \nabla L(\theta_{t-1}) \\
             &= \theta_{t - 1} - \eta D^{-1} \big(H \theta_{t - 1} \big) \\
             &= \big( I - \eta D^{-1} H \big) \theta_{t - 1} \\
             &= \big( I - \eta D^{-1} H \big)^t \theta_{0} 
\end{align*} 

As such, this simple model would suggest that the max eigenvalue of the following matrix may be related to training instability of models trained with Adam
\begin{align} \label{eqn:stability_cond}
    \lambda_{max}(D^{-1} H ) = \lambda_{max}(D^{-1/2} H D^{-1/2}).
\end{align}

While \eqref{eqn:stability_cond} does not take into the account the effects of adaptive preconditioning or momentum, we find some empirical evidence that this approximation provides understanding into the stability of the optimization. 

\begin{figure}[h]
     \centering
     \includegraphics[width=\textwidth]{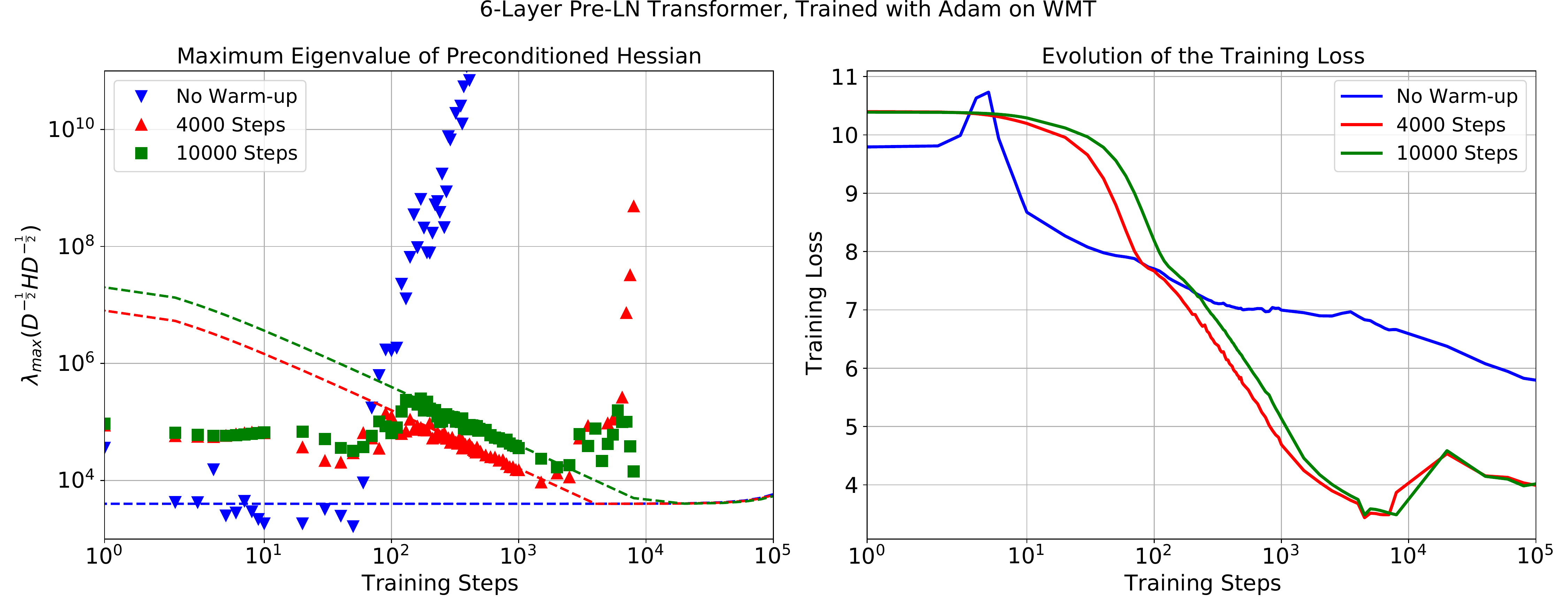}
     \caption{Instabilities in model training are reflected in the loss curvature even for models trained with Adam. Left: Maximum eigenvalues of the preconditioned Hessian throughout the training. Dashed line correspond to $40/\eta$. Right: Evolution of the training loss for different models. Training becomes unstable exactly when the eigenvalues cross the threshold. \label{fig:adam}}
\end{figure}

Figure \ref{fig:adam} below examines the evolution of $\lambda_{max}(D^{-1/2} H D^{-1/2})$ for three Transformer models trained with Adam and different warm-up lengths. Here, $D$ is a diagonal matrix with $D_{i, i} = \sqrt{\mbox{Corrected Adam grad squared EMA}} + \mbox{Adam } \epsilon$. We observe that, similar to the models trained with momentum, the maximum (preconditioned) Hessian eigenvalue adapts to the warm-up schedule (green and red markers). We notice that --perhaps due to the effect of momentum or adaptive preconditioning -- the threshold $2 / \eta$ does not seem to be aligning with the data well. Instead, an empirically corrected threshold $40 / \eta$ seems to fit the data better. We observe that instabilities in model training coincide exactly with $\lambda_{max}(D^{-1/2} H D^{-1/2})$ crossing the empirically corrected threshold.

These observations suggest that some of the insights discussed in the main text seem to carry over to the case of adaptive optimizers. We leave further exploration of this more complex setting to future work.

\begin{figure}[h]
     \centering

    \includegraphics[width=\textwidth]{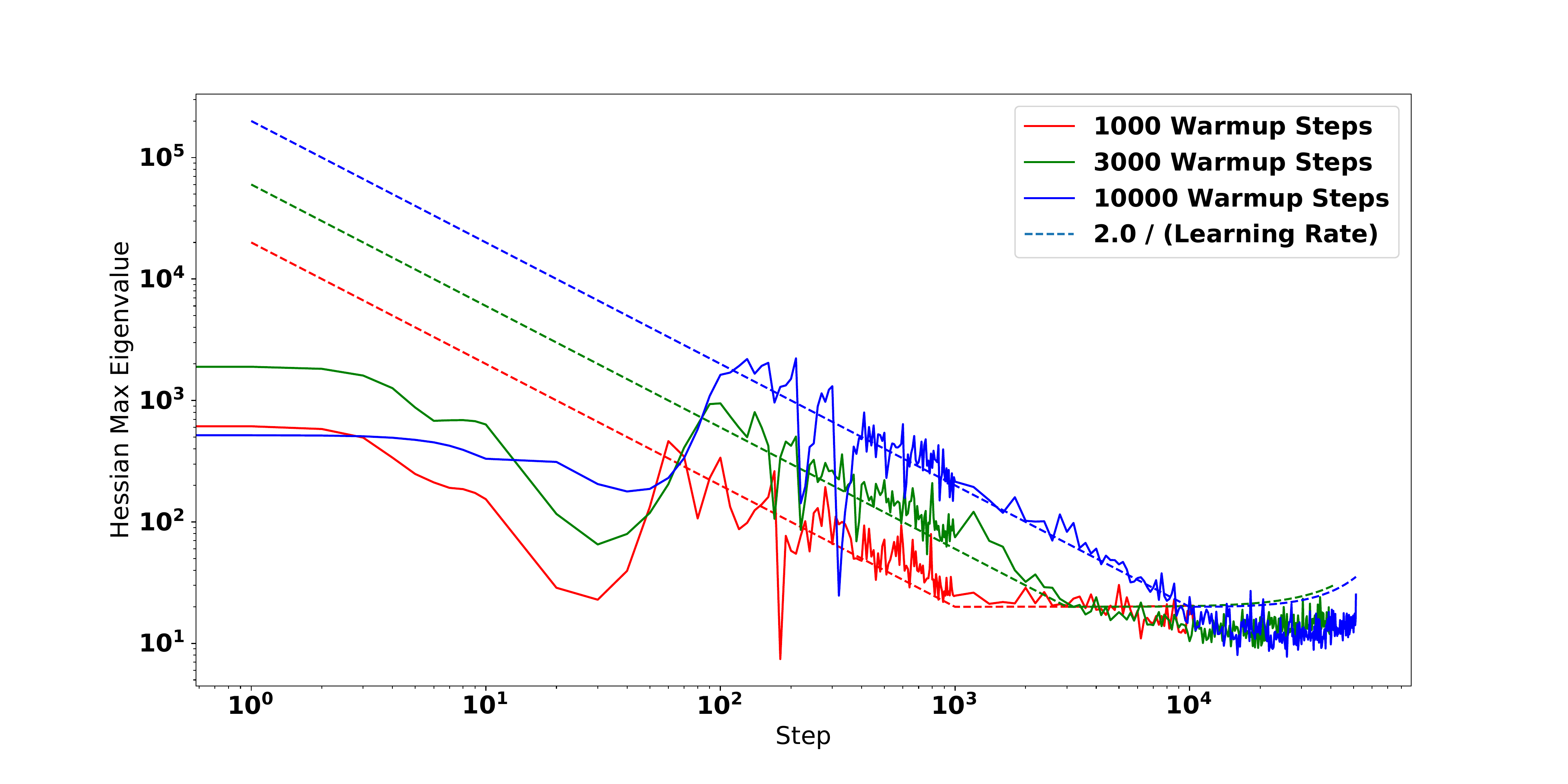}

     \caption{Further evidence of learning rate warmup ``pushing'' the optimization trajectory towards regions with reduced $\lambda_1$. Note the rate at which $\lambda_1$ changes closely matches depends on the length of the warmup. Model shown is the non-BN WideResnet (standard init) trained at batch size 2048 with peak learning rate of $.1$.}
     \label{fig:wrn_warmup_steps}
\end{figure}

\begin{figure}[h]
     \centering

    \includegraphics[width=\textwidth]{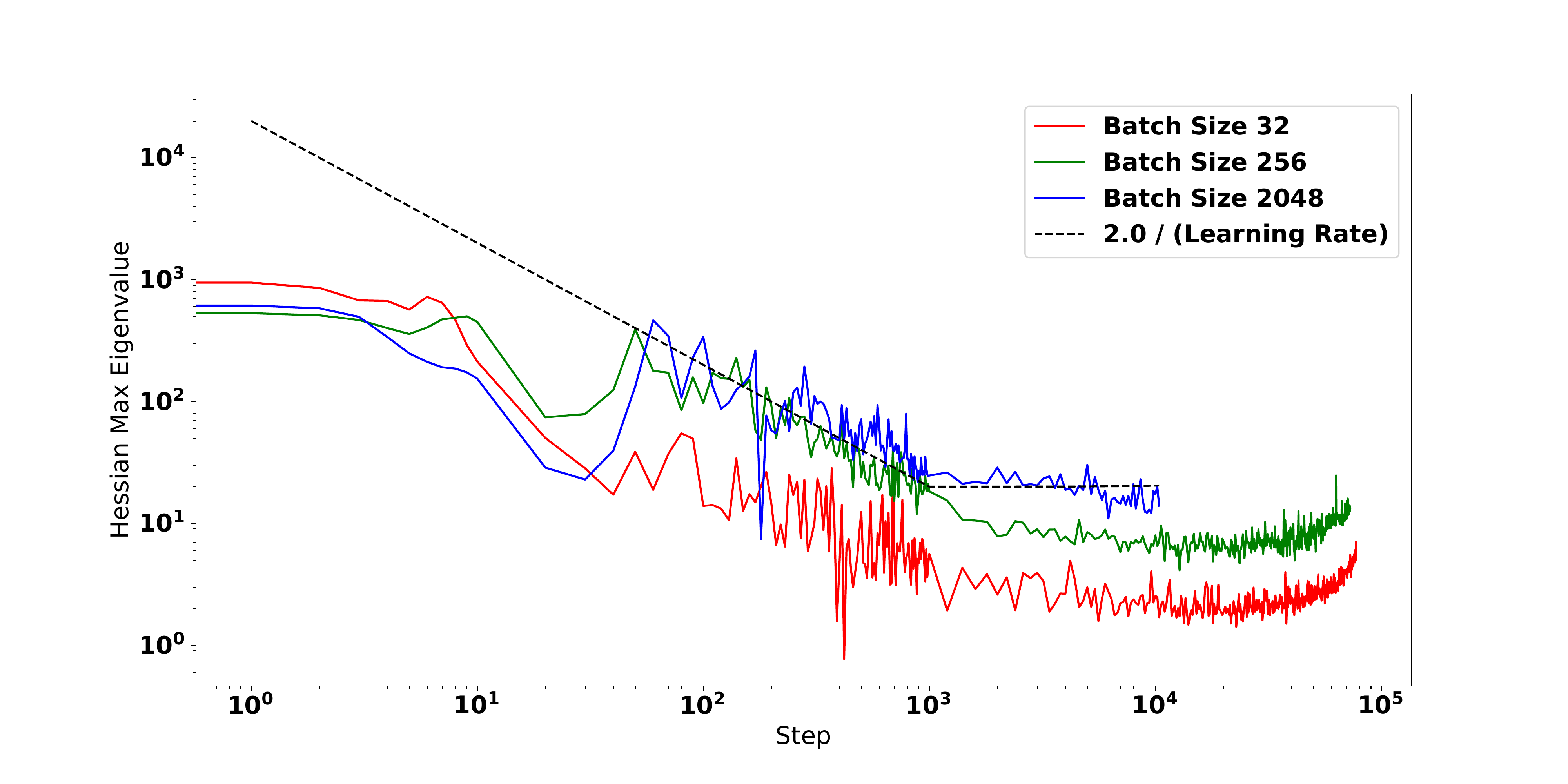}

     \caption{\citet{wu2018sgd} predicts that the stability bound should increase with smaller values of the gradient variance $\Sigma$ during training. This plot confirms this prediction by increasing the batch size. For a small batch size ($32$), the mid training curvature hovers significantly below the $2.0 / \eta$ approximation. As the batch size increases to $2048$, the mid training curvature is larger, approaching the $2.0 / \eta$ approximation. All curves show the WRN without batch norm, trained with the same learning schedule using 1000 steps of warmup.}
     \label{fig:wrn_bs_warmup}
\end{figure}

\clearpage
\section{Compute Resources Used}

Nearly all experiments utilized the Google Cloud Platform with v2 cloud TPUs except for the following: The Figure 2 Resnet-50 and Densenet experiments utilized the v3 cloud TPU, while the GradInit code was run on a cloud machine with a single V100 GPU. The Figure 2 experiments were done in parallel using up to 50 v2 TPUs concurrently over the period of a few days. Additionally, all the Machine Translation models were trained on v3 cloud TPUs.

\end{document}